\documentclass[10pt,journal,compsoc]{IEEEtran}

\ifCLASSOPTIONcompsoc
  \usepackage[nocompress]{cite}
\else
  \usepackage{cite}
\fi

\ifCLASSINFOpdf
\else
\fi
\usepackage{array}
\usepackage{marvosym}
\usepackage{tabularx}
\usepackage{amsmath}
\usepackage{graphicx}
\usepackage{algorithmic}
\usepackage{threeparttable}
\usepackage{titletoc}
\usepackage[ruled,linesnumbered]{algorithm2e}
\usepackage{cite}
\usepackage{amsfonts}
\usepackage{color}
\usepackage{xcolor}
\usepackage{soul}
\usepackage{colortbl}

\usepackage{bm}
\usepackage{enumitem}

\usepackage{bbding}
\usepackage{multirow}
\usepackage{indentfirst}
\usepackage{hyperref}   
\usepackage{color,xcolor}
\definecolor{mygray}{gray}{.9}
\definecolor{dark-red}{rgb}{0.6, 0.1, 0.1}
\definecolor{dark-green}{rgb}{0, 0.6, 0.25} 
\definecolor{citecolor}{rgb}{0,0.443,0.737} 
\definecolor{linkcolor}{rgb}{0.956,0.298,0.235}

\definecolor{clsdatasetcolor}{HTML}{dd4e70}
\definecolor{segdatasetcolor}{HTML}{76d1e6}
\definecolor{progdatasetcolor}{HTML}{39cf60}
\definecolor{rgdatasetcolor}{HTML}{e39b77}
\definecolor{voclsdatasetcolor}{HTML}{b379bd}
\definecolor{retdatasetcolor}{HTML}{b0c515}
\definecolor{vqadatasetcolor}{HTML}{848ff8}

\definecolor{bestresultcolor}{HTML}{BFE1C9}
\definecolor{subresultcolor}{HTML}{FEFAC2}

\definecolor{yellowarrow}{HTML}{ECCF66}

\hypersetup{
    colorlinks=true,
    linkcolor=linkcolor,
    filecolor=magenta,
    citecolor=dark-green,
    urlcolor=dark-green 
}

\newcommand{\black}[1]{{\color{black}#1}}
\usepackage{booktabs}
\usepackage{caption}
\begin{document}

\title{ASAP: Advancing medical volumetric representation learning with anatomy-aware semantically-adaptive pre-training}
\newcommand{\algname}{ASAP\xspace}
\author{Rongsheng Wang, Fenghe Tang, Zihang Jiang, Yingtai Li, Xu Zhang, Haoran Lai, Wenxin Ma, Wei Wei, Zhiyang He, Xiaodong Tao, Rui Yan, Qingsong Yao, and Shaohua Kevin Zhou,~\IEEEmembership{Fellow,~IEEE}
        
\IEEEcompsocitemizethanks{
\IEEEcompsocthanksitem Rongsheng Wang, Fenghe Tang, Zihang Jiang, Yingtai Li, Xu Zhang, Haoran Lai, Wenxin Ma, Rui Yan, and Shaohua Kevin Zhou are with School of Biomedical Engineering, Division of Life Sciences and Medicine, University of Science and Technology of China (USTC), Hefei, Anhui, China 230026; Medical Imaging, Robotics, Analytic Computing \& Learning (MIRACLE) Lab, YRD-RIGHT, USTC Suzhou Institute for Advanced Research, Suzhou, Jiangsu, China 215123; Jiangsu Provincial Key Laboratory of Multimodal Digital Twin Technology, Suzhou, Jiangsu, China 215123; and also with Biomedical Basic Research Center (BBRC) of Jiangsu Province, Suzhou, Jiangsu, China 215123.
(E-mail: \{\href{mailto:rongsheng_wang@mail.ustc.edu.cn}{\black{rongsheng\_wang@mail.ustc.edu.cn}}\})
\IEEEcompsocthanksitem Wei Wei is with Department of Radiology, The First Affiliated Hospital of USTC, Division of Life
Sciences and Medicine, USTC, Hefei, Anhui, 230001, China.
\IEEEcompsocthanksitem Zhiyang He and Xiaodong Tao are with Anhui IFLYTEK CO., Ltd.
\IEEEcompsocthanksitem Qingsong Yao is with the School of Medicine, Stanford University.
\IEEEcompsocthanksitem Shaohua Kevin Zhou is also with State Key Laboratory of Precision and Intelligent Chemistry, Hefei, Anhui, China 230026.
\IEEEcompsocthanksitem{This paper is an extension of our MICCAI 2025 paper\cite{wang2025simcrop}.}
\IEEEcompsocthanksitem{Corresponding author: Shaohua Kevin Zhou (E-mail: \{\href{mailto:skevinzhou@ustc.edu.cn}{\black{skevinzhou@ustc.edu.cn}}\}). Co-corresponding author: Qingsong Yao (E-mail: \{\href{mailto:qingsongyao98@gmail.com}{\black{qingsongyao98@gmail.com}}\}).}
}
}


\IEEEtitleabstractindextext{%
\begin{abstract}

Learning transferable and interpretable representations from medical volumetric scans remains challenging due to complex anatomical structures and weak, heterogeneous supervision provided by radiology reports. 
In this paper, we propose \textbf{A}natomy-aware \textbf{S}emantically-\textbf{A}daptive \textbf{P}re-training (\algname), a principled vision–language pre-training framework for fine-grained medical volumetric representation learning from large-scale chest CT scans and their corresponding radiology reports.
\algname integrates three key components: (1) an anatomy-aware knowledge injection module that incorporates organ-level structural priors via off-the-shelf segmentation tool to encourage anatomically coherent representations; (2) a semantically-adaptive selective alignment mechanism that dynamically associates sentence-level findings with localized volumetric regions; and (3) a semantically-adaptive fusion module for effective interaction between anatomically informed visual features and grounded textual cues under dual-modal masked modeling paradigm. 
Beyond methodological contributions, we establish a comprehensive benchmark for medical volumetric vision--language pre-training on chest CT, covering 15 datasets and 22 downstream tasks spanning abnormality classification, segmentation, disease prognosis prediction, report generation, vocabulary classification, cross-modal retrieval and visual question answering. 
This benchmark provides standardized evaluation protocols to systematically assess representation quality under diverse clinical settings and data regimes. 
Extensive experiments demonstrate that \algname consistently achieves state-of-the-art performance across tasks and datasets, with particularly pronounced gains under limited supervision and distribution shift, validating its effectiveness in learning transferable and clinically meaningful volumetric representations. 
Codes, models, benchmark are available at \href{https://github.com/ToniChopp/ASAP}{https://github.com/ToniChopp/ASAP}.

\end{abstract}

}

\maketitle
\IEEEdisplaynontitleabstractindextext
\IEEEpeerreviewmaketitle

\ifCLASSOPTIONcompsoc
\IEEEraisesectionheading{\section{Introduction}\label{sec:intro}}
\else
\section{Introduction}
\label{sec:intro}
\fi

\IEEEPARstart{D}{eep} learning (DL) has demonstrated remarkable efficacy in medical image representation learning~\cite{xie2024unimiss+,azad2024medsegreview,li2025visionunite,yan2026pamt,zhou2021review}, yielding expert-level performance across a wide range of diagnostic tasks when trained with large-scale, high-quality annotations~\cite{wu2023minimizingseg,pathak2022deep,chen2022review,zhao2022diagnoselikeradiologist,he2025homeomorphism}. 
However, the heavy reliance on curated supervision remains a fundamental obstacle to scalability and generalization in medical imaging~\cite{zhou2022referes,zhang2023kad,wu2025radfm,zhou2023pcrlv2,liu2023medmllm}. 
This limitation is particularly pronounced in 3D volumetric medical imaging, where the annotation cost from doctors scales dramatically with spatial resolution and slice count, making voxel-wise or region-level labeling expensive and difficult to scale~\cite{ma2022abdomenct,zhang2024pmcvqa,zheng2024large,niu2025m3fm,tang2026hiendmae}. 
Consequently, reducing dependence on manual annotation has become a central challenge for advancing volumetric medical representation learning. 

Medical vision-language pre-training (Med-VLP) offers a promising alternative by leveraging radiological reports as naturally occurring semantic supervision~\cite{zhang2022convirt,huang2021gloria,wang2022mgca,wang-etal-2022-medclip,wu2023medklip}. 
Extensive studies on 2D radiographs have shown that aligning images with free-text reports enables models to learn clinically meaningful representations without explicit disease labels~\cite{tiu2022chexzero,zhou2023mrm,li2024mlip,huang2024maco,Wang2025ECAMP,lian2025alta}. 
These results establish Med-VLP as a powerful paradigm for mitigating annotation scarcity and exploiting the rich semantic information embedded in radiological narratives. 
Despite its success in 2D settings, extending Med-VLP to 3D volumetric medical imaging presents substantially greater challenges. 

\begin{figure*}[t!]
    \centering
    \includegraphics[width=\linewidth]{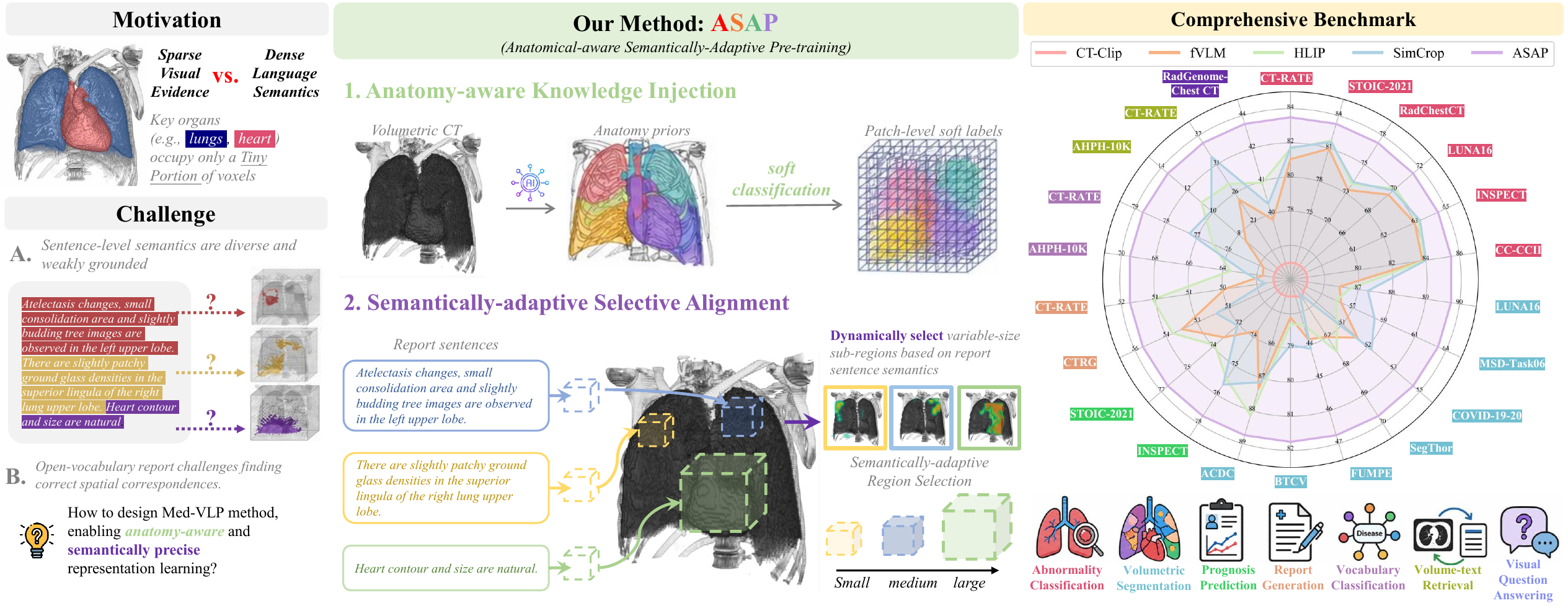}
    \caption{\textbf{(Left) Motivation:} Existing 3D Med-VLP methods are limited by spatial sparsity of informative anatomical structures and the heterogeneous, open-vocabulary nature of radiology reports, leading to weak and noisy cross-modal alignment. \textbf{(Middle) Method:} \algname addresses these challenges via (1) anatomy-aware knowledge injection, which introduces organ-level priors as patch-level soft supervision, and (2) semantically-adaptive selective alignment, which grounds sentence-level semantics to volumetric regions with adaptive spatial extent. \textbf{(Right) Benchmark:} A comprehensive benchmark to evaluate 3D Med-VLP methods spanning 7 downstream categories and 22 tasks demonstrates that \algname consistently outperforms prior methods, yielding more transferable and clinically relevant volumetric representations.}
    \vspace{-5mm}
    \label{fig:teaser}
\end{figure*}

Unlike 2D radiographs, 3D volumetric scans exhibit severe spatial sparsity and structural imbalance~\cite{cao2024biud,blankemeier2026merlin,wu2025vocolarge}. The clinically relevant anatomical regions in 3D volume only occupy a small fraction of the overall volume, while the majority corresponds to non-informative background. As illustrated in Fig.~\ref{fig:teaser}, key organs such as the lungs, heart, and aorta account for only a limited proportion of voxels in chest CT scans. 
Under such conditions, global or coarse-grained cross-modal objectives tend to be dominated by background regions, leading to diluted supervision and suboptimal representation learning. 

More fundamentally, the difficulty of volumetric Med-VLP lies not only in spatial sparsity but also in the nature of disease semantics. 
Diseases in 3D medical imaging do not form a closed or finite category set; instead, they exhibit diverse manifestations, variable spatial extents, and open-ended linguistic descriptions in reports~\cite{hamamci2024ctclip,draelos2021radchestct,vaya2020bimcv}.
Radiological reports are semantically dense and often describe findings at the sentence or phrase level, covering both global impressions and localized anatomical observations. 
However, these descriptions are weakly grounded in space and display heterogeneous semantic granularity, ranging from high-level summaries to fine-grained regional cues~\cite{fvlm_iclr25,cao2025vsd-boost}.
{\em This intrinsic mismatch between sparse 3D visual evidence and semantically rich but weakly localized textual supervision poses a critical challenge for volumetric Med-VLP}. 

Existing volumetric Med-VLP approaches~\cite{hamamci2024ctclip,lin2024ctglip} primarily rely on global alignment or coarse cross-modal interactions, which are insufficient to establish fine-grained, sentence-conditioned visual grounding under severe spatial sparsity. 
As a result, learned representations may capture superficial semantic correlations while failing to actively identify disease-relevant regions and their corresponding textual descriptions. 
Therefore, a key open problem is how to design volumetric Med-VLP frameworks that can encourage sentence-conditioned localized correspondences between sparse 3D anatomical structures and open-vocabulary radiological reports, enabling anatomy-aware and semantically precise representation learning. 

To address the above challenges, we propose \textbf{A}natomy-aware \textbf{S}emantically-\textbf{A}daptive \textbf{P}re-training ({\bf \algname}), a unified framework for localized and precise 3D medical volumetric representation learning. 
As illustrated in Fig.~\ref{fig:teaser}, \algname is designed to systematically exploit two complementary sources of supervision that are naturally available in volumetric medical imaging: 
\textbf{semantically rich but spatially implicit radiological reports}, and 
\textbf{structural anatomical priors inherent in 3D volumetric scans}. 
\algname is built upon a dual-modal masked modeling paradigm that serves as a common pre-training backbone, enabling joint encoding of volumetric images and reports into a shared representation space. 
Within this unified framework, we introduce three tightly coupled, task-specific modules that explicitly address the unique challenges of volumetric Med-VLP: (i) \textit{anatomy-aware knowledge injection}, (ii) \textit{semantically-adaptive selective alignment}, and (iii) \textit{semantically-adaptive fusion}.

Different from existing approaches for 3D volumetric representation learning, which typically treat volumetric patches uniformly and overlook explicit anatomical organization~\cite{hamamci2024ctclip,bai2024m3d,Liu2025T3D,Park_2025_radzero3d}, \algname incorporates organ-level structural priors as an integral component of representation learning. 
Specifically, we exploit off-the-shelf anatomical segmentation tools~\cite{zhao2025sat,wasserthal2023totalsegmentator} to obtain organ-level structural cues at scale, without introducing additional annotation cost. 
Based on these cues, the proposed \textit{anatomy-aware knowledge injection} module injects compact and precise anatomy-aware knowledge into the latent space, enabling the model to capture anatomically coherent spatial layouts and structural dependencies under severe volumetric sparsity. 

Beyond anatomical structure, effectively grounding radiological language in volumetric visual evidence remains a central challenge for Med-VLP. 
Radiological reports, particularly the \textit{``Findings''} section, are semantically rich but spatially implicit, describing a potentially unbounded set of disease patterns without explicit localization. 
Existing approaches~\cite{fvlm_iclr25,cao2025vsd-boost,zhao2025hlip} often rely on predefined organ- or disease-level categories, modeling anatomical structure and pathology semantics in a largely decoupled manner. 
In contrast, \algname introduces a \textit{semantically-adaptive selective alignment} mechanism that adaptively associates descriptive sentences with informative volumetric sub-regions based on their semantic relevance.
By conditioning visual correspondence on sentence-level semantics, this module enables the vision encoder to acquire open-vocabulary disease knowledge and establish fine-grained, sentence-conditioned visual grounding. 
As a result, rich and diverse disease semantics expressed in radiological reports can be gracefully mapped onto a finite set of anatomically meaningful visual sub-regions, facilitating effective cross-modal representation learning beyond global alignment.

Finally, effectively exploiting fine-grained cross-modal correspondences during report reconstruction remains non-trivial.
Most existing mask-based cross-modal reconstruction frameworks rely on either simple feature aggregation~\cite{zhou2023mrm,lian2025alta} or standard cross-attention mechanisms~\cite{chen2022m3ae}, where visual tokens are treated uniformly during decoding. 
Such designs are agnostic to the semantic relevance of visual regions and may dilute sentence-specific cues, especially in volumetric settings characterized by severe spatial sparsity and heterogeneous semantic granularity. 
In contrast, \algname explicitly leverages the fine-grained, sentence-conditioned visual representations learned by the semantically-adaptive selective alignment module. 
We introduce a \textit{semantically-adaptive fusion} mechanism that injects semantics-embedding visual evidence into the cross-attention process, guiding report reconstruction to attend preferentially to clinically salient and semantically grounded anatomical regions. 
By conditioning cross-modal fusion on aligned visual priors, this module enables more effective utilization of localized visual evidence and promotes coherent integration of global anatomical context and sentence-level semantics.

To comprehensively assess the effectiveness and generalizability of volumetric Med-VLP models, we further establish a large-scale evaluation benchmark for medical vision--language pre-training. 
In contrast to prior studies~\cite{cao2024biud,fvlm_iclr25,lin2024ctglip,hamamci2024ctclip,cao2025vsd-boost,zhao2025hlip,bai2024m3d,Liu2025T3D}, which are typically evaluated on a limited set of datasets or task configurations, our benchmark systematically covers 22 downstream tasks under multiple training data regimes, together with four external validation tasks.
These tasks span a diverse range of clinical scenarios, including disease classification, organ and lesion segmentation, prognosis prediction, report generation, open-vocabulary disease classification, volume--report retrieval, and visual question answering. 
Extensive experimental results across this benchmark consistently demonstrate that the proposed \algname yields substantial performance improvements over existing Med-VLP approaches, establishing new state-of-the-art results on the majority of evaluated tasks.
These results validate the effectiveness of our semantically-adaptive and anatomy-aware pre-training strategy in learning transferable and clinically meaningful representations for 3D medical volumetric imaging. 

This work substantially extends our preliminary study by introducing both methodological and experimental advances tailored for volumetric Med-VLP. 
The main extensions are summarized as follows:
\begin{itemize}
    \item \textbf{Anatomy-aware knowledge injection with organ-level priors.}
    We introduce explicit organ-level structural knowledge to enforce anatomically coherent representations under severe volumetric sparsity.
    \item \textbf{Semantically-adaptive selective alignment.}
    We generalize the original alignment to a semantically-adaptive formulation that dynamically selects sentence-conditioned volumetric sub-regions for fine-grained visual grounding.
    \item \textbf{Semantically-adaptive fusion.}
    We replace fixed cross-attention and global pooling with a semantic-aware fusion mechanism that emphasizes clinically relevant visual evidence during report reconstruction.
    \item \textbf{Expanded and rigorous evaluation.}
    The evaluation is extended from a limited task set to a comprehensive benchmark with 22 downstream tasks under multiple data regimes and four external validations.
\end{itemize}

\section{Related Work}
\subsection{Large-scale Vision--Language Pre-training}
Vision--Language Pre-training (VLP) has become a fundamental paradigm for learning transferable multi-modal representations from large-scale paired image--text data~\cite{clip,jia2021ALIGN,coca,li2022glip,li2022flip}, playing a critical role in the development of general-purpose multi-modal foundation models~\cite{alayrac2022flamingo,li2023blipv2,liu2023llava}. 
The core objective of VLP is to devise scalable paradigms that effectively distill semantic correlations from weakly aligned vision--language supervision, thereby minimizing the dependency on explicit manual annotations. 
Existing methodologies can be categorized based on their pre-training objectives and interaction mechanisms. 

\noindent\textbf{Contrastive Representation Learning.} Pioneering works, such as ALIGN~\cite{jia2021ALIGN}, CLIP~\cite{clip}, and their variants~\cite{li2022glip,li2022flip,zhai2023siglip}, primarily adopt a dual-encoder architecture coupled with contrastive learning objectives. By aligning visual and textual representations in a shared global hyperspace, these methods demonstrate remarkable zero-shot transferability. However, constrained by the lack of deep cross-modal interaction layers, such paradigms often struggle to capture intricate, fine-grained structural correspondences and lack the capability for generative reasoning. 

\noindent\textbf{Masked Multi-modal Modeling.} To facilitate deep and dense cross-modal fusion, another stream of research aggregates visual patches and linguistic tokens into a unified sequence, employing Masked Signal Modeling (MSM) to predict and reconstruct corrupted tokens conditioned on context~\cite{lu2019vilbert,su2020vlbert,bao2021beit,Singh2022flava}. Representative frameworks like BEiT-3~\cite{wang2023beit3} leverage this objective to learn robust, modality-agnostic representations, effectively modeling the joint probability distribution of image and text features. 

\noindent\textbf{Generative and Hybrid Frameworks.} In parallel, generative paradigms, exemplified by the BLIP family~\cite{li2023blipv2}, have shifted focus towards Language Modeling (LM) objectives. By treating text generation as the primary supervision while incorporating auxiliary alignment tasks, frameworks like CoCa~\cite{coca} and Flamingo~\cite{alayrac2022flamingo} couple representation learning with conditional generation, enabling sophisticated semantic composition and open-ended reasoning capabilities. 

Collectively, these advancements provide a theoretical blueprint for multi-modal representation learning, demonstrating that robust features can be derived from noisy, weakly supervised data~\cite{liu2025valor,zhang2025unsupervised}. 
This success naturally motivates the translation of VLP paradigms to domain-specific scenarios, particularly medical imaging, where vast archives of paired diagnostic scans and expert-authored radiology reports are available but underutilized~\cite{johnson2019mimic,vaya2020bimcv,hamamci2024ctclip}. However, the distinct domain characteristics of medical imaging—ranging from subtle pathologies in 2D radiographs to complex anatomical structures in 3D volumes—necessitate tailored adaptations of these general methodologies. 
In the following section, we review the evolution of Med-VLP frameworks from 2D projectional imaging to more challenging 3D volumetric modalities.

\vspace{-3mm}
\subsection{Medical Vision--Language Pre-training}
Medical vision--language pre-training (Med-VLP) has emerged as an effective paradigm for alleviating the reliance on manual annotations in medical imaging by leveraging paired images and radiological reports~\cite{xie2024unimiss+,zhou2022referes,zhang2023kad,tiu2022chexzero,yao2025evax}. 
Existing studies have demonstrated consistent improvements in representation learning, particularly for 2D chest radiographs, validating the effectiveness of vision--language supervision in capturing clinically meaningful semantics without dense annotations~\cite{Wang2025ECAMP,wang2022mgca,zhou2023mrm,wan2023medUniC,XIE2024MedIM}. 
More recently, attention has shifted toward extending Med-VLP to 3D volumetric imaging, motivated by the richer anatomical and pathological information encoded in CT and MRI scans~\cite{wu2025radfm,zhang2024pmcvqa,wu2024voco,wu2025vocolarge,zhuang2025mim}. 
However, the transition from 2D to 3D Med-VLP remains challenging due to the increased spatial complexity and the weakly localized nature of textual supervision in volumetric settings.

\vspace{-3mm}
\subsubsection{Med-VLP for 2D Radiographs}
Early Med-VLP studies for 2D radiographs mainly adopt contrastive learning to align visual representations with textual descriptions, enabling effective global semantic alignment from paired chest X-rays and reports~\cite{zhang2022convirt,tiu2022chexzero,zhou2022referes,wang-etal-2022-medclip,huang2024enhancing}. 
Although local contrastive variants such as Med-UniC~\cite{wan2023medUniC}, Imitate~\cite{liu2024Imitate} and MGVLA~\cite{yan2025mgvla} have been proposed, contrastive supervision remains biased toward coarse-grained alignment, limiting its capacity to model fine-grained cross-modal interactions. 
Subsequent works including MLIP~\cite{li2024mlip}, and SAHN~\cite{li2025SAHN} further combine knowledge guidance and hard-negative mining, improving discriminability within this paradigm.

Beyond contrastive alignment, masked reconstruction-based frameworks explicitly model local cross-modal correspondences by jointly reconstructing masked visual and textual tokens~\cite{chen2022m3ae,zhou2023mrm}, enabling finer-grained representation learning under the relatively simple spatial structure of 2D radiographs by encouraging token-level reasoning. 
Representative methods such as MedIM~\cite{XIE2024MedIM} and ECAMP~\cite{Wang2025ECAMP} further enrich masked modeling with comprehensive masking strategies and explicit clinical context semantics.

More recently, hybrid paradigms integrate contrastive alignment with reconstruction or context construction objectives to unify global semantic alignment and local correspondence modeling. 
Methods such as PRIOR~\cite{cheng2023prior}, MaCo~\cite{huang2024maco}, G2D~\cite{liu2024g2d}, and ALTA~\cite{lian2025alta} jointly optimize alignment and reconstruction-based objectives, leading to more expressive cross-modal representations.

Overall, 2D Med-VLP methods benefit from dense visual semantics and relatively limited spatial complexity. 
Extending these paradigms to 3D volumetric imaging, however, poses fundamentally different challenges under weakly localized supervision and complex anatomical structures.

\subsubsection{Med-VLP for 3D Volumetric Imaging}
In contrast to the abundance of 2D radiographs, acquiring large-scale 3D medical cohorts (e.g., CT, MRI) is impeded by higher acquisition costs, privacy constraints, and radiation risks~\cite{ma2022abdomenct,withers2021xray,he2023gsvl,zhuang2025mim,niu2025m3fm}. 
Consequently, owing to this data scarcity, early explorations in 3D Med-VLP predominantly adopted contrastive learning paradigms to distill discriminative features from limited paired data.

Pioneering works, such as CT-CLIP~\cite{hamamci2024ctclip} and M3D~\cite{bai2024m3d}, straightforwardly extend 2D contrastive frameworks to 3D by performing global alignment between volumetric embeddings and report tokens. 
To mitigate the loss of local details inherent in global pooling, subsequent studies inject inductive biases to refine cross-modal alignment.
Specifically, approaches like BIUD~\cite{cao2024biud} leverage cross-modal knowledge distillation from 2D pre-trained experts. 
UniMedI~\cite{he2024UniMedI} constructs unified semantic spaces to extract volumetric representations. 
Furthermore, recent hierarchical methods, such as fVLM~\cite{fvlm_iclr25}, ViSD-Boost~\cite{cao2025vsd-boost}, MG-3D~\cite{ni2024mg3d}, HLIP~\cite{zhao2025hlip} and MedVista3D~\cite{li2025medvista3d}, incorporate anatomical priors to align sub-volumes with corresponding textual phrases at multiple granularities.
While these strategies improve upon global baselines, they remain bound by the fundamental limitations of discriminative objectives: they assume static, pre-defined correspondences between modalities and lack the generative capacity to model the underlying distribution of complex anatomy.

Notwithstanding these improvements, a critical methodological discrepancy persists: the intrinsic mismatch between the high-dimensional, spatially sparse nature of 3D volumetric imaging and the semantic density of radiology reports.
Unlike 2D projectional images, 3D volumetric scans exhibit severe \textbf{spatial redundancy}, where informative pathologies occupy a negligible fraction of the total volume.
Standard contrastive objectives, which optimize global or coarse-regional similarity, are susceptible to being dominated by background noise, failing to effectively ground fine-grained textual findings (e.g., small nodules) to their precise voxel-level locations.
This issue is exacerbated by the lack of massive datasets, preventing models from implicitly learning these localized features through scale alone.

Departing from discriminative paradigms that treat alignment as a static matching problem, we reformulate 3D Med-VLP as a \textbf{generative masked modeling} task.
By synergizing Anatomy-aware Knowledge Injection and Semantically-Adaptive Alignment within a dual-modal masked auto-encoding framework, we explicitly address the challenges of anatomical sparsity and granularity mismatch.
Unlike prior efforts~\cite{hamamci2024ctclip,bai2024m3d,Liu2025T3D,fvlm_iclr25,cao2025vsd-boost,ni2024mg3d,zhao2025hlip,li2025medvista3d} that address these issues in isolation, our framework \textit{proactively} exploits textual semantics to guide the discovery of sparse visual features, enabling fine-grained, anatomy-aware representation learning even under limited volumetric supervision.

\begin{figure*}[t!]
    \centering
    \includegraphics[width=\linewidth]{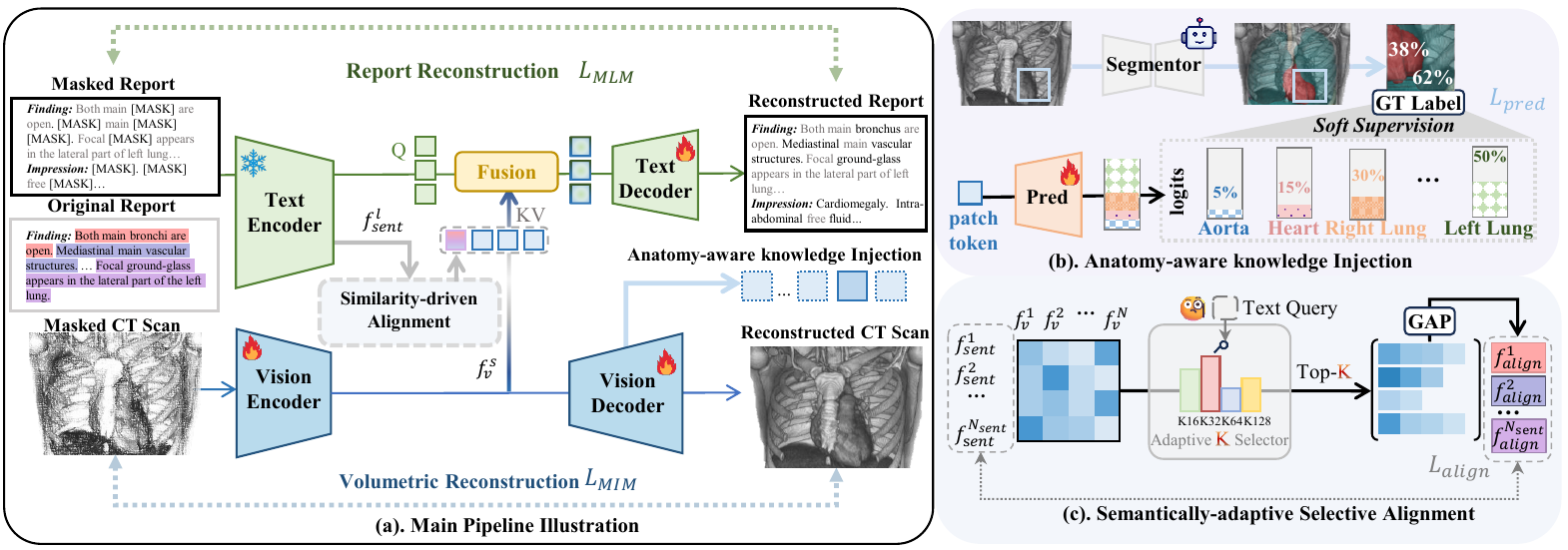}
    \caption{Overview of the proposed \algname framework.
    \textbf{(a).} \algname adopts a dual-modal masked modeling framework for joint learning from volumetric CT scans and radiology reports. Given masked inputs, visual and textual encoders extract volumetric and sentence-level representations. A \textit{semantically-adaptive selective alignment (SSA)} module establishes sentence-conditioned correspondences via similarity-driven region selection, followed by a \textit{semantically-adaptive fusion (Fusion)} module for report reconstruction, while a vision decoder reconstructs the masked volume.  
    \textbf{(b).} \textit{Anatomy-aware knowledge injection (AKI)} incorporates organ-level priors from off-the-shelf segmentation as patch-level soft supervision on intermediate features, encouraging anatomically consistent representations. 
    \textbf{(c).} For each sentence, SSA adaptively selects top-$K$ relevant patches based on cross-modal similarity, enabling grounding at variable spatial extents and reducing background-dominated alignment.}
    \vspace{-5mm}
    \label{fig:framework}
\end{figure*}

\section{Method} 
In this section, we elaborate on the methodology of \algname, a framework designed for fine-grained chest CT representation learning via large-scale vision-language pre-training on volumetric scans and corresponding radiology reports. 
As illustrated in Fig.~\ref{fig:framework}, \algname is built upon a dual-modal masked modeling paradigm that jointly encodes volumetric visual patterns and radiological language semantics (Sec.~\ref{sec:mask}).
Building upon this shared masked representation backbone, \algname incorporates three anatomically and semantically informed learning components to progressively enhance representation quality:
\textbf{Anatomy-aware knowledge injection} (Sec.~\ref{sec:anatomy_injection}), which injects anatomical constraints through a fine-grained, structure-aware prediction objective;
\textbf{Semantically-adaptive selective alignment} (Sec.~\ref{sec:similar}), which establishes sentence-specific correspondences between radiological findings and localized volumetric sub-regions;
\textbf{Semantically-adaptive fusion} (Sec.~\ref{sec:fusion}), which further refines the aligned visual--textual representations via smooth and adaptive integration guided by structural and linguistic priors.
By holistically coupling these components, \algname facilitates the learning of comprehensive, interpretable, and clinically meaningful chest CT representations at a fine-grained level.

\vspace{-3mm}
\subsection{Dual-Modal Masked Modeling}
\label{sec:mask}
Our approach is built upon a dual-modal masked autoencoder paradigm~\cite{chen2022m3ae}, which establishes a foundational representation space by jointly modeling volumetric CT scans and unstructured radiology reports through complementary self-supervised objectives. Following prior medical vision language pre-training studies~\cite{bai2024m3d,cao2024biud,fvlm_iclr25,zhao2025hlip}, we employ a vision transformer (ViT)~\cite{vit} and a BERT~\cite{devlin2019bert} as the vision and text encoder, respectively. 

\noindent\textbf{Volumetric masked modeling.} Given a volumetric CT scan $I \in \mathbb{R}^{1\times H\times W\times D}$, we first partition it into $N_p$ non-overlapping 3D patches $\{I^s\}_{s=1}^{N_p}$. Following the masked autoencoder strategy~\cite{He2021MAE}, a random masking operator $\mathcal{M}_v(\cdot)$ is applied to the patch sequence, where each patch is independently masked with probability $\alpha$. This results in a subset of $N_u = (1-\alpha)N_p$ visible patches, denoted as $I_u=\{I_u^s\}_{s=1}^{N_u}$. Each visible patch is augmented with a 3D positional embedding $E^{v}_{\rm pos}$ constructed using sinusoidal encoding~\cite{zhou2023self}, and then processed by the vision encoder $E_v$ to obtain latent volumetric representations: $f_v=E_v(I_u+E^{v}_{\rm pos})$. During decoding, the latent features $f_v$ are concatenated with a set of learnable mask tokens $f_m$ at the masked patch locations and fed into the volumetric decoder $D_v$ to reconstruct the full volume: $\hat{I}=D_v(f_v, f_m)$. The reconstruction objective is formulated as a mean squared error (MSE) loss between the reconstructed volume and the original input: 
\begin{equation}
    \mathcal{L}_{\rm {MIM}}(\hat{I}, I) = \mathbb{E}_{I \sim \mathcal{P}_v}\!\left[ \| \hat{I} - I \|_2^2 \right],
\end{equation}
where $\mathcal{P}_v$ denotes the empirical distribution of volumetric CT scans in the pre-training dataset.

\noindent\textbf{Radiology report masking.} Given a radiology report $T$ consisting of $N_t$ words, we apply a token-level masking strategy analogous to masked language modeling~\cite{devlin2019bert}. Specifically, a random masking operator $\mathcal{M}_t(\cdot)$ masks each token with probability $\gamma$, producing a set of masked tokens $T_m=\{T_m^l\}_{l=1}^{N_m}$ and unmasked tokens $T_u=\{T_u^l\}_{l=1}^{N_t-N_m}$. Both masked and unmasked tokens are jointly processed by the text encoder $E_t$ to generate contextualized report representations: $f_{t}=E_t(T_m, T_u)$. Unlike conventional masked language modeling, we do not directly reconstruct the masked tokens. Instead, the masked textual representations serve as semantic anchors for subsequent cross-modal interaction and fusion, which will be detailed in Sec.~\ref{sec:fusion}.

This dual-modal masked modeling paradigm establishes modality-specific yet semantically compatible latent spaces, forming the foundation for the proposed semantically-adaptive alignment.

\begin{table}[t]
\small
\centering
\setlength{\tabcolsep}{ 5pt}
\caption{Summary of the 18 downstream datasets included in our evaluation benchmark, comprising 15 datasets for internal evaluation and 3 datasets for external validation. The symbol $\dagger$ denotes our privately collected datasets, and $\ddagger$ indicates datasets used exclusively for external validation.}
\begin{tabular}{lll}
\toprule
Dataset  &Modality  &Task\\
\midrule
Rad-ChestCT~\cite{draelos2021radchestct}  &CT  &\textcolor{clsdatasetcolor}{Lung Abnormality Cls.}\\
\rowcolor[gray]{1.0}CC-CCII~\cite{he2020cc-ccii}  &CT  &\textcolor{clsdatasetcolor}{COVID \& Pneumonia Cls.}\\
MSD-Lung~\cite{antonelli2022msd}  &CT  &\textcolor{segdatasetcolor}{Lung Tumor Seg.}\\
\rowcolor[gray]{1.0}COVID-19-20~\cite{roth2022covid-19-20}  &CT  &\textcolor{segdatasetcolor}{COVID Seg.}\\
SegThor~\cite{lambert2020segthor}  &CT  &\textcolor{segdatasetcolor}{Thoracic Risk Seg.}\\
\rowcolor[gray]{1.0}FUMPE~\cite{masoudi2018FUMPE}  &CT  &\textcolor{segdatasetcolor}{Pulmonary Embolism Seg.}\\
BTCV~\cite{landman2015btcv}  &CT  &\textcolor{segdatasetcolor}{Abdomen Seg.}\\
\rowcolor[gray]{1.0}ACDC~\cite{bernard2018acdc}  &MRI  &\textcolor{segdatasetcolor}{Heart Seg.}\\
CTRG~\cite{tang2024ctrg}  &CT \& VL  &\textcolor{rgdatasetcolor}{Report Generation}\\
\rowcolor[gray]{1.0}{CT-Rate~\cite{hamamci2024ctclip}}  &{CT \& VL}  &\textcolor{clsdatasetcolor}{Lung Abnormality Cls.}\\
\rowcolor[gray]{1.0}&  &\textcolor{rgdatasetcolor}{Report Generation}\\
\rowcolor[gray]{1.0}&  &\textcolor{voclsdatasetcolor}{Vocabulary Cls.}\\
\rowcolor[gray]{1.0}&  &\textcolor{retdatasetcolor}{Volume-report Retrieval}\\
{LUNA16~\cite{setio2017luna16}}  &{CT}  &\textcolor{clsdatasetcolor}{Lung Nodule Cls.}\\
&  &\textcolor{segdatasetcolor}{Lung Seg.}\\
\rowcolor[gray]{1.0}{INSPECT~\cite{huang2023inspect}}  &{CT}  &\textcolor{clsdatasetcolor}{Pulmonary Embolism Cls.}\\
\rowcolor[gray]{1.0}&  &\textcolor{progdatasetcolor}{Prognosis Prediction}\\
{STOIC-2021~\cite{revel2021stoic}}  &{CT}  &\textcolor{clsdatasetcolor}{COVID Cls.}\\
&  &\textcolor{progdatasetcolor}{Prognosis Prediction}\\
RadGenome~\cite{zhang2025radgenomect}  &{CT \& VL}  &\textcolor{vqadatasetcolor}{Visual Question Answering}\\
\rowcolor[gray]{1.0}{AHPH-10K $\dagger$}  &{CT \& VL}  &\textcolor{voclsdatasetcolor}{Vocabulary Cls.}\\
\rowcolor[gray]{1.0}&  &\textcolor{retdatasetcolor}{Volume-report Retrieval}\\
\rowcolor[gray]{1.0}&  &\textcolor{clsdatasetcolor}{Lung Abnormality Cls.}$\ddagger$\\\midrule
RSPECT~\cite{colak2021rspect}$\ddagger$  &CT  &\textcolor{clsdatasetcolor}{Pulmonary Embolism Cls.}\\
\rowcolor[gray]{1.0}RIDER~\cite{aerts2014rider}$\ddagger$  &CT  &\textcolor{segdatasetcolor}{Lung Tumor Seg.}\\
COVID-19-  &\multirow{2}{*}{CT}  &\textcolor{segdatasetcolor}{Lung Seg.}\\
-CT-Seg~\cite{Ma2021COVID-19-SegBenchmark}$\ddagger$&  &\textcolor{segdatasetcolor}{COVID Seg.}\\
\bottomrule
\end{tabular}
\vspace{-5mm}
\label{Talbe:dataset}
\end{table}

\subsection{Anatomy-Aware Knowledge Injection}
\label{sec:anatomy_injection}
Anatomical structure priors introduce precise organ-level semantics in a compact and finite supervision space, enabling precise structural guidance that effectively complements large-scale self-supervised pre-training~\cite{lin2024ctglip,fvlm_iclr25,cao2025vsd-boost}.
To leverage such priors, we propose a fine-grained anatomical structure prediction objective that injects organ-level constraints into volumetric representations in a patch-wise manner.
Given an input volumetric scan $I \in \mathbb{R}^{1\times H\times W\times D}$, we define a finite set of $N_o$ anatomical organs $\mathcal{O}=\{O^{(1)}, O^{(2)}, \dots, O^{(N_o)}\}$ as critical structural categories for pre-training.   
We employ an off-the-shelf segmentation tool~\cite{wasserthal2023totalsegmentator}, to automatically generate the corresponding voxel-wise anatomical mask $M_{\rm seg} \in \mathbb{R}^{H\times W\times D}$, which provides scalable anatomy-aware knowledge without additional annotation cost. 
As illustrated in Fig.~\ref{fig:framework} (b), this anatomical information is incorporated as an auxiliary self-supervised objective during volumetric masked modeling.

Practically, during the decoding stage of volumetric masked modeling, we extract intermediate decoder features rather than the final reconstruction output.
These intermediate representations capture sufficient anatomical structure while avoiding over-reliance on low-level appearance details, making them well suited for knowledge supervision. 
To construct patch-level anatomical labels, we first partition the segmentation mask $M_{\rm seg}$ into $N_p$ non-overlapping 3D patches $\{M_{\rm seg}^s\}_{s=1}^{N_p}$, consistent with the patchification scheme used for original volumetric scan. For each patch $M_{\rm seg}^s$ and anatomical organ $O^{(i)}$, we compute the proportion of voxels belonging to that organ within the patch:
\begin{equation}
y_{s,i} = \frac{1}{|M_{\rm seg}^s|} \sum_{v \in M_{\rm seg}^s} \mathbb{I}\big(v \in O^{(i)}\big),
\end{equation}
where $\mathbb{I}(\cdot)$ denotes the indicator function and $y_{s,i} \in [0,1]$ represents the soft anatomical label. This formulation naturally captures partial organ occupancy within a patch, providing more fine-grained and spatially faithful supervision.

Let $f_{\rm dec}^s$ denote the intermediate decoder feature corresponding to the $s$-th patch. We predict organ presence scores $\hat{y}_{s,i}$ for each anatomical category via a lightweight prediction head.
The anatomy-aware structure prediction loss is then formulated as a multi-label binary cross-entropy (BCE) loss:
\begin{equation}
\mathcal{L}_{\rm pred} = \frac{1}{N_p} \sum_{s=1}^{N_p} \sum_{i=1}^{N_o} \Big[-y_{s,i} \log \hat{y}_{s,i} - (1 - y_{s,i}) \log (1 - \hat{y}_{s,i}) \Big].
\end{equation}
By supervising patch-level representations with soft anatomical proportions, the proposed objective encourages the model to encode anatomically coherent and structurally discriminative features, while remaining robust to structural ambiguity and partial volume effects.

\vspace{-3mm}
\subsection{Semantically-Adaptive Selective Alignment}
\label{sec:similar}
Given a radiology report $T=[T_{\rm F}, T_{\rm I}]$, where $T_{\rm F}$ and $T_{\rm I}$ denote the ``\textit{Finding}" and ``\textit{Impression}" sections respectively, we focus on the ``\textit{Finding}" section due to its clinical relevance in describing localized visual evidence within volumetric scans. Unlike the ``\textit{Impression}" section, which primarily provides high-level diagnostic summarization, the ``\textit{Finding}" section contains fine-grained, spatially grounded observations that are well suited for fine-grained cross-modal alignment. 

Specifically, for a "\textit{Finding}" section comprising $N_{\rm sent}$ linguistically independent sentences, we tokenize each sentence and feed the resulting sentence set $T_{\rm sent}=\{T_{\rm sent}^l\}_{l=1}^{N_{\rm sent}}$ into the text encoder $E_t$ to derive sentence-level semantic representations: $f_{\rm sent}=E_t(T_{\rm sent})$. 
As illustrated in Fig.~\ref{fig:framework} (c), we establish fine-grained semantic correspondences between textual findings and localized volumetric structures by computing the cosine similarity between sentence-level representations $f_{\rm sent}=\{f_{\rm sent}^l\}_{l=1}^{N_{\rm sent}}$ and latent volumetric patch features $f_v=\{f_v^s\}_{s=1}^{N_u}$. 
Specifically, the similarity between the $l$-th finding sentence and the $s$-th unmasked volumetric patch is defined as:
\begin{equation}
\begin{aligned}
    \mathrm{Sim}_{l,s} = \frac{(f_{\rm sent}^l)^{\top} f_v^s} {\|f_{\rm sent}^l\|_2 \, \|f_v^s\|_2},
    l \in \{1,\dots,N_{\rm sent}\},
    s \in \{1,\dots,N_u\}.
\end{aligned}
\end{equation}
This similarity score quantitatively reflects the semantic correspondence between radiological descriptions and localized volumetric regions. 

\noindent\textbf{Semantically-adaptive patch selection.}
For each ``\textit{Finding}" sentence, a subset of spatially relevant volumetric patches is selected for representation alignment. A straightforward strategy is to select a fixed number of top-$K$ patches with the highest similarity scores. However, radiological findings inherently exhibit varying semantic granularity: some sentences describe diffuse or global patterns (e.g., extensive ground-glass opacities), whereas others correspond to highly localized abnormalities (e.g., a pulmonary nodule). 
Enforcing a fixed $K$ across all sentences may therefore lead to suboptimal alignment, either introducing irrelevant regions or failing to capture sufficient contextual evidence. 
To address this issue, we introduce a semantically-adaptive patch selection mechanism that dynamically determines the number of aligned volumetric patches conditioned on sentence semantics. As depicted in Fig.~\ref{fig:framework} (c), the adaptive selector predicts a sentence-specific patch budget $K_l$ based on the semantic representation of the $l$-th sentence. 
Formally, given the sentence-level semantic representation $f_{\rm sent}^l$, we predict a latent scalar score:
\begin{equation}
    z_l = g_{\theta}(f_{\rm sent}^l),
\end{equation}
where $g_{\theta}(\cdot)$ denotes a lightweight learnable predictor that estimates the semantic spatial extent associated with the $l$-th finding sentence.
To ensure numerical stability and boundedness, the latent score is mapped through a sigmoid activation:
\begin{equation}
\tilde{z}_l = \sigma(z_l), \quad \tilde{z}_l \in (0,1).
\end{equation}
We then discretize $\tilde{z}_l$ into a finite set of predefined multi-scale candidates $\mathcal{K} = \{K^{(1)}, K^{(2)}, \dots, K^{(M)}\}$ with $M = |\mathcal{K}|$. Specifically, we compute the index:
\begin{equation}
    m_l = \left\lfloor M \cdot \tilde{z}_l \right\rfloor + 1,
\end{equation}
and selecting the corresponding patch number $K_l = \mathcal{K}[m_l]$. Since the selector only predicts a discrete patch budget rather than directly participating in gradient-based patch optimization, the discretization operation does not impede end-to-end training. 
This adaptive selection mechanism enables each finding sentence to align with an appropriate spatial extent, effectively translating continuous semantic relevance into discrete and interpretable volumetric scales.
By restricting $K_l$ to a finite candidate set, the proposed selector avoids degenerate solutions while preserving sufficient flexibility to capture cross-granularity semantic variations.

\noindent\textbf{Selective alignment.}
Given the semantically-adaptive patch number $K_l$, we identify the most semantically relevant volumetric regions by selecting a subset of patches that maximizes aggregated similarity under a cardinality constraint. Formally, the index set of selected patches for the $l$-th sentence is defined as:
\begin{equation}
\mathcal{S}_l =
\arg\max_{\mathcal{S} \subseteq \{1,\dots,N_u\}, \, |\mathcal{S}| = K_l}
\sum_{s \in \mathcal{S}} \mathrm{Sim}_{l,s}.
\end{equation}
This adaptive selection enables each finding sentence to attend to a spatial extent commensurate with its semantic granularity, avoiding both over- and under-alignment induced by a fixed patch budget.
Given the adaptive index set $\mathcal{S}_l$ for the $l$-th finding sentence, we extract the corresponding subset of volumetric patch features $\{f_v^s\}_{s \in \mathcal{S}_l}$. These discriminative visual features are spatially aggregated via global average pooling (GAP) to obtain a sentence-aligned visual representation:
\begin{equation}
f_{\rm align}^l = \frac{1}{|\mathcal{S}_l|} \sum_{s \in \mathcal{S}_l} f_v^s .
\end{equation}
To enforce semantic coherence between sentence-level textual representations and their corresponding localized visual evidence, we adopt a similarity-driven fine-grained contrastive learning objective. 
Specifically, we employ a symmetric InfoNCE loss~\cite{Oord2018RepresentationLW} defined over sentence-aligned visual features $\{f_{\rm align}^l\}$ and sentence embeddings $\{f_{\rm sent}^l\}$:
\begin{equation}
\begin{aligned}
\mathcal{L}_{\rm align}
= -\frac{1}{N_{\rm sent}} \sum_{i=1}^{N_{\rm sent}}
\Bigg[
& \log \frac{\exp(\mathrm{sim}(f_{\rm align}^i, f_{\rm sent}^i)/\tau)}
{\sum_{j=1}^{N_{\rm sent}} \exp(\mathrm{sim}(f_{\rm align}^i, f_{\rm sent}^j)/\tau)} \\
+\,\,
& \log \frac{\exp(\mathrm{sim}(f_{\rm sent}^i, f_{\rm align}^i)/\tau)}
{\sum_{j=1}^{N_{\rm sent}} \exp(\mathrm{sim}(f_{\rm sent}^i, f_{\rm align}^j)/\tau)}
\Bigg],
\end{aligned}
\end{equation}
where $\mathrm{sim}(\cdot,\cdot)$ denotes the cosine similarity between $\ell_2$-normalized features, and $\tau$ is the temperature hyperparameter set to $0.07$ following common practice. 
By aligning sentence-level semantics with adaptively aggregated local visual evidence, the proposed objective promotes fine-grained yet semantically consistent cross-modal representation learning.

\vspace{-3mm}
\subsection{Semantically-adaptive Fusion}
\label{sec:fusion}
Different from conventional multi-modal frameworks that rely on standard cross-attention~\cite{chen2022m3ae,liu2023m3ae} or direct feature summation~\cite{zhou2023mrm}, we propose a semantically-adaptive fusion mechanism that explicitly conditions cross-attention on sentence-aligned visual evidence. 
Instead of heuristically modulating attention weights, we inject the learned fine-grained alignment representations as explicit semantic priors into the visual latent space, enabling the decoder to attend to anatomically and semantically salient regions during report reconstruction. 

\noindent\textbf{Semantic-augmented localized fusion:} Given the latent volumetric patch features $\{f_v^s\}_{s=1}^{N_u}$ and the sentence-aligned visual representations $\{f_{\rm align}^l\}_{l=1}^{N_{\rm sent}}$ obtained from Sec.~\ref{sec:similar}, we construct a semantic-aware similarity token that summarizes sentence-conditioned visual evidence.
Specifically, we aggregate the sentence-aligned features into a compact semantic prior:
\begin{equation}
f_{\rm sim}
=
\gamma \cdot
\frac{1}{N_{\rm sent}}
\sum_{l=1}^{N_{\rm sent}}
f_{\rm align}^{l},
\end{equation}
where $\gamma$ is a learnable scaling factor. 
The semantic-aware similarity token is then prepended to the volumetric patch sequence, forming a semantic-augmented localized visual latent representation:
\begin{equation}
\tilde{f}_v
=
\{ f_{\rm sim} \}
\cup
\{ f_v^{1}, \dots, f_v^{N_u} \}.
\end{equation}
This semantic-augmented sequence serves as the key–value memory for cross-modal attention, allowing the decoder to jointly reason over global semantic cues and localized anatomical structures. 

Given masked textual representations $\{f_t^l\}_{l=1}^{N_t}$, we apply a standard cross-attention operation, where masked text tokens act as queries and the augmented visual latent sequence $\tilde{f}_v$ serves as keys and values.
This design enables masked tokens to attend not only to local volumetric patches but also to the injected semantic-aware similarity token, which provides sentence-conditioned global context. 

\noindent\textbf{Global fusion:} In parallel, we extract a global visual context vector via global average pooling over volumetric patch features: $g_v = \frac{1}{N_u} \sum_{s=1}^{N_u} f_v^s$ via global average pooling over the volumetric patch features, capturing holistic anatomical information.
The final fused representation for each masked text token is obtained by:
\begin{equation}
\tilde{{f}}_t^{l} = \mathrm{LayerNorm}(c_t^{l} + g_v),
\end{equation}
where we adopt element-wise addition (broadcasting $g_v$) followed by normalization for efficient fusion.
The fused representations $\{\tilde{f}_t^{l}\}$ are subsequently fed into the text decoder $D_t$ to reconstruct the masked report tokens.
Formally, the masked language modeling objective is defined as
\begin{equation}
\mathcal{L}_{\mathrm{MLM}}=-\frac{1}{N_m}\sum_{l=1}^{N_m}\log P\left(T_m^{l}\mid D_t(\tilde{f}_t^{l}) \right),
\end{equation}
where $T_m^{l}$ denotes the ground-truth masked token and $N_m$ is the number of masked positions.

Finally, the overall training objective of \algname is formulated as
\begin{equation}
\mathcal{L}
=
\mathcal{L}_{\mathrm{MIM}}
+
\lambda_1 \mathcal{L}_{\mathrm{pred}}
+
\lambda_2 \mathcal{L}_{\mathrm{align}}
+
\lambda_3 \mathcal{L}_{\mathrm{MLM}}.
\end{equation}
Unless otherwise specified, we empirically set $\lambda_1=\lambda_2=\lambda_3=1$ to balance reconstruction, prediction and cross-modal alignment objectives.

\begin{table*}[t]
\small
\centering
\setlength{\tabcolsep}{2.0pt}
\caption{Performance comparison on six medical volumetric classification benchmarks under linear probing settings.
We report AUC (\%) and ACC (\%). Except for methods with $^*$, all methods are evaluated using a 3D ViT-B~\cite{vit} backbone. 
The best and second-best results are highlighted with \colorbox{bestresultcolor}{first} and \colorbox{subresultcolor}{second}, respectively.
${}^{\boldsymbol{\dagger}}$ denotes that the proposed method significantly outperforms the second-best approach under the same setting, based on a paired Wilcoxon signed-rank test ($p < 0.05$).}
\begin{tabular}{l|ccc|ccc|cc|cc|cc|ccc}
\toprule
\multirow{3}{*}{Method}  &\multicolumn{3}{c|}{CT-RATE}  &\multicolumn{3}{c|}{STOIC-2021}  &\multicolumn{2}{c|}{Rad-ChestCT}  &\multicolumn{2}{c|}{LUNA16}  &\multicolumn{2}{c|}{INSPECT}  &\multicolumn{3}{c}{CC-CCII}\\ 
&\multicolumn{3}{c|}{(AUC)}  &\multicolumn{3}{c|}{(AUC)}  &\multicolumn{2}{c|}{(AUC)}  &\multicolumn{2}{c|}{(AUC)}  &\multicolumn{2}{c|}{(AUC)}  &\multicolumn{3}{c}{(ACC)}\\
&\multicolumn{1}{c}{1\%} &\multicolumn{1}{c}{10\%} &\multicolumn{1}{c|}{100\%} &\multicolumn{1}{c}{1\%} &\multicolumn{1}{c}{10\%} &\multicolumn{1}{c|}{100\%}  &\multicolumn{1}{c}{10\%} &\multicolumn{1}{c|}{100\%} &\multicolumn{1}{c}{10\%} &\multicolumn{1}{c|}{100\%}  &\multicolumn{1}{c}{10\%} &\multicolumn{1}{c|}{100\%}  &\multicolumn{1}{c}{1\%} &\multicolumn{1}{c}{10\%} &\multicolumn{1}{c}{100\%}\\\midrule                  
Random init                      &$58.2_{0.4}$  &$60.9_{0.3}$  &$63.8_{0.0}$  &$61.8_{0.1}$  &$62.0_{0.1}$  &$62.3_{0.1}$  &$56.4_{2.3}$  &$60.0_{1.5}$  &$57.6_{1.3}$  &$60.4_{0.8}$  &$51.8_{0.6}$  &$55.2_{0.1}$  &$62.9_{1.1}$  &$77.3_{0.4}$  &$82.2_{0.4}$\\\midrule
\multicolumn{3}{l}{\textit{\textcolor{gray}{3D Med-SSL}}}\\
MAE~\cite{He2021MAE}             &$75.6_{0.5}$  &$78.1_{0.3}$  &$79.9_{0.0}$  &$64.6_{0.2}$  &$71.6_{0.4}$  &$75.6_{0.3}$  &$68.6_{0.9}$  &$71.8_{0.6}$  &$66.3_{1.2}$  &$70.5_{0.2}$  &$60.1_{0.3}$  &$63.1_{0.1}$  &$71.0_{0.6}$  &$83.4_{0.7}$  &$87.8_{0.1}$\\
PCRLv2~\cite{zhou2023pcrlv2}  &$77.1_{0.3}$  &$81.3_{0.2}$  &$81.9_{0.0}$  &$67.2_{0.3}$  &$73.1_{0.2}$  &$76.1_{0.4}$  &$69.8_{0.6}$  &$72.3_{0.3}$  &$66.7_{0.9}$  &$70.8_{0.2}$  &$60.3_{0.4}$  &$63.0_{0.1}$  &$73.6_{0.5}$  &$84.1_{0.5}$  &$88.8_{0.2}$\\
HEMAE~\cite{tang2026hiendmae}  &$76.7_{0.1}$  &$78.9_{0.2}$  &$80.6_{0.0}$  &$66.2_{0.1}$  &$73.4_{0.2}$  &$75.9_{0.2}$  &$69.4_{0.5}$  &$71.4_{0.2}$  &$67.0_{0.2}$  &$72.7_{0.1}$  &$61.2_{0.1}$  &$63.0_{0.1}$  &$72.8_{0.4}$  &$83.7_{0.4}$  &$88.1_{0.1}$\\\midrule
\multicolumn{3}{l}{\textit{\textcolor{gray}{3D Med-VLP}}}\\
M3AE~\cite{chen2022m3ae}         &$77.2_{0.4}$  &$79.5_{0.2}$  &$81.0_{0.0}$  &$65.0_{0.4}$  &$72.8_{0.2}$  &$76.5_{0.6}$  &$69.3_{0.7}$  &$72.2_{0.3}$  &$67.1_{0.3}$  &$70.8_{0.1}$  &$60.6_{0.1}$  &$63.3_{0.3}$  &$73.2_{0.4}$  &$84.0_{0.4}$  &$88.5_{0.7}$\\
MRM~\cite{zhou2023mrm}           &$77.7_{0.2}$  &$81.7_{0.1}$  &$82.1_{0.0}$  &$74.6_{0.1}$  &$78.4_{0.4}$  &$80.2_{0.0}$  &$72.1_{0.5}$  &$72.6_{0.4}$  &$66.6_{0.8}$  &$72.2_{0.2}$  &$61.3_{0.3}$  &$63.6_{0.2}$  &$74.1_{1.3}$  &$82.2_{0.5}$  &$89.6_{0.1}$\\
CT-CLIP~\cite{hamamci2024ctclip} &$74.1_{1.0}$  &$78.6_{0.6}$  &$80.4_{0.0}$  &$70.0_{0.3}$  &$74.3_{0.5}$  &$77.5_{0.4}$  &$67.5_{1.0}$  &$69.1_{0.8}$  &$64.0_{1.4}$  &$65.0_{0.5}$  &$59.4_{0.5}$  &$62.5_{0.1}$  &$70.7_{0.9}$  &$80.3_{0.4}$  &$86.4_{0.4}$\\
M3D~\cite{bai2024m3d}            &$73.9_{0.7}$  &$79.1_{0.5}$  &$80.7_{0.0}$  &$70.5_{0.4}$  &$74.5_{0.3}$  &$77.9_{0.2}$  &$68.0_{1.5}$  &$69.8_{0.9}$  &$65.2_{1.1}$  &$68.4_{0.9}$  &$60.0_{0.6}$  &$62.2_{0.1}$  &$71.7_{0.4}$  &$80.6_{0.7}$  &$87.2_{0.6}$\\
Merlin$^*$~\cite{blankemeier2026merlin} &$75.8_{0.4}$  &$78.3_{0.1}$  &$79.6_{0.0}$  &$68.6_{0.6}$  &$71.4_{0.3}$  &$73.1_{0.2}$  &$70.2_{0.2}$  &$73.8_{0.2}$  &$65.4_{0.2}$  &$68.5_{0.3}$  &$62.4_{0.2}$  &$65.2_{0.1}$  &\cellcolor{subresultcolor}$77.2_{0.2}$  &$85.2_{0.8}$  &$90.5_{0.1}$\\
fVLM~\cite{fvlm_iclr25}          &$80.2_{0.2}$  &$82.1_{0.1}$  &$82.7_{0.0}$  &$78.6_{0.5}$  &$82.2_{0.3}$  &$85.0_{0.1}$  &$72.9_{0.4}$  &$75.1_{0.3}$  &$66.8_{0.5}$  &$72.9_{0.2}$  &\cellcolor{subresultcolor}$62.8_{0.4}$  &$65.1_{0.2}$  &$76.1_{0.9}$  &\cellcolor{subresultcolor}$86.1_{0.3}$  &$91.3_{0.2}$\\
ViSDB$^*$~\cite{cao2025vsd-boost} &$80.5_{0.1}$  &$82.0_{0.2}$  &$82.4_{0.0}$  &$78.2_{0.5}$  &$81.7_{0.3}$  &$84.7_{0.1}$  &$73.1_{0.3}$  &$74.7_{0.3}$  &$67.0_{0.2}$  &$73.4_{0.2}$  &$62.0_{0.3}$  &$64.7_{0.1}$  &$75.8_{0.4}$  &$86.0_{0.4}$  &$91.0_{0.3}$\\
HLIP~\cite{zhao2025hlip}         &$80.7_{0.1}$  &\cellcolor{subresultcolor}$82.5_{0.1}$  &\cellcolor{subresultcolor}$83.0_{0.0}$  &\cellcolor{subresultcolor}$79.0_{0.3}$  &$83.0_{0.2}$  &\cellcolor{subresultcolor}$85.5_{0.1}$  &\cellcolor{subresultcolor}$73.4_{0.5}$  &$75.5_{0.4}$  &\cellcolor{subresultcolor}$67.1_{0.2}$  &\cellcolor{subresultcolor}$73.5_{0.1}$  &$62.7_{0.2}$  &$65.4_{0.2}$  &$76.5_{0.4}$  &$86.0_{0.5}$  &\cellcolor{subresultcolor}$91.4_{0.4}$\\
SimCroP~\cite{wang2025simcrop}   &\cellcolor{subresultcolor}$81.0_{0.1}$  &$82.4_{0.1}$  &$82.9_{0.0}$  &$78.8_{0.2}$  &\cellcolor{subresultcolor}$83.2_{0.2}$  &$85.2_{0.0}$  &\cellcolor{subresultcolor}$73.4_{0.5}$  &\cellcolor{subresultcolor}$75.8_{0.2}$  &$67.0_{0.4}$  &$73.3_{0.2}$  &$62.7_{0.3}$  &\cellcolor{subresultcolor}$65.5_{0.3}$  &$76.8_{0.3}$  &$85.9_{0.5}$  &$91.3_{0.2}$\\
\rowcolor[gray]{1.0} \algname (Ours)  &\cellcolor{bestresultcolor}${82.2_{0.1}^{\boldsymbol{\dagger}}}$  &\cellcolor{bestresultcolor}${83.5_{0.0}^{\boldsymbol{\dagger}}}$  &\cellcolor{bestresultcolor}${84.1_{0.0}^{\boldsymbol{\dagger}}}$  &\cellcolor{bestresultcolor}${81.2_{0.5}^{\boldsymbol{\dagger}}}$  &\cellcolor{bestresultcolor}${83.7_{0.0}^{\boldsymbol{\dagger}}}$  &\cellcolor{bestresultcolor}${86.1_{0.0}^{\boldsymbol{\dagger}}}$  &\cellcolor{bestresultcolor}${76.1_{0.1}^{\boldsymbol{\dagger}}}$  &\cellcolor{bestresultcolor}${79.0_{0.0}^{\boldsymbol{\dagger}}}$  &\cellcolor{bestresultcolor}${68.5_{0.1}^{\boldsymbol{\dagger}}}$ &\cellcolor{bestresultcolor}${74.3_{0.2}^{\boldsymbol{\dagger}}}$  &\cellcolor{bestresultcolor}${63.3_{0.1}}$  &\cellcolor{bestresultcolor}${65.9_{0.3}}$  &\cellcolor{bestresultcolor}${77.9_{0.2}^{\boldsymbol{\dagger}}}$  &\cellcolor{bestresultcolor}${86.8_{0.2}^{\boldsymbol{\dagger}}}$  &\cellcolor{bestresultcolor}${92.8_{0.1}^{\boldsymbol{\dagger}}}$\\\bottomrule
\end{tabular}
\vspace{-2mm}
\label{Table:LinearProbe}
\end{table*}

\section{Experiments}
\subsection{Dataset and Implementation Details}
\noindent\textbf{Pre-training datasets.}
We conduct medical vision--language pre-training on CT-RATE~\cite{hamamci2024ctclip}, currently the largest publicly available dataset of paired CT volumetric scans and radiological reports, comprising 50,188 scan--report pairs. 
Following the official data split, 47,149 volumetric scan--report pairs from the training subset are used for pre-training. 
For \textit{anatomy-aware knowledge injection}, we employ TotalSegmentator~\cite{wasserthal2023totalsegmentator} to generate organ-level segmentation masks as pseudo anatomical supervision. 
Consistent with prior work~\cite{fvlm_iclr25}, we retain a subset of nine clinically relevant thoracic organs to provide compact and reliable anatomical priors to balance anatomical coverage with segmentation reliability in chest CT scans\footnote{The selected organs include: \textit{left upper lung lobe}, \textit{left lower lung lobe}, \textit{right upper lung lobe}, \textit{right middle lung lobe}, \textit{right lower lung lobe}, \textit{heart}, \textit{left atrial appendage}, \textit{esophagus}, and \textit{aorta}.}.

\noindent\textbf{Evaluation benchmark. }
We establish a comprehensive evaluation benchmark designed to assess Med-VLP methods on chest CT scan--report pairs. 
As summarized in Table~\ref{Talbe:dataset}, the benchmark comprises 18 downstream datasets spanning diverse task types, including abnormality classification, segmentation, prognosis prediction, report generation, vocabulary classification, and volume--report retrieval. 
Among them, 16 datasets are publicly available, and one private dataset, AHPH-10K\footnote{AHPH-10K is collected from Anhui Provincial Hospital (The First Affiliated Hospital of USTC) between 2021 and 2023, containing 10,000 paired chest CT scans and radiological reports. Abnormality labels are extracted from reports using Qwen2.5-VL~\cite{Qwen2.5-VL}}, is included to support both internal and external validation under realistic clinical domain shifts. 
For CT-RATE~\cite{hamamci2024ctclip}, Rad-ChestCT~\cite{draelos2021radchestct}, LUNA16~\cite{setio2017luna16}, CC-CCII~\cite{he2020cc-ccii}, MSD-Lung~\cite{antonelli2022msd}, COVID-19-20~\cite{roth2022covid-19-20}, SegThor~\cite{lambert2020segthor}, FUMPE~\cite{masoudi2018FUMPE}, BTCV~\cite{landman2015btcv}, ACDC~\cite{bernard2018acdc}, and CTRG~\cite{tang2024ctrg}, we follow the standard training/validation/testing splits adopted in prior studies~\cite{wang2025simcrop,wu2025vocolarge} to ensure fair comparison. 
For INSPECT~\cite{huang2023inspect}, we use the official data split provided by the dataset, while for STOIC-2021~\cite{revel2021stoic}, we randomly partition the data into 80\% for training and 20\% for testing. 
To systematically assess the quality of volumetric representations learned during pre-training, we evaluate all methods under multiple labeled-data regimes (1\%, 10\%, and 100\% of the training data) for abnormality classification, segmentation, and prognosis prediction tasks.
In addition, external validation is conducted on AHPH-10K, RSPECT~\cite{colak2021rspect}, RIDER~\cite{aerts2014rider}, and COVID-19-CT-Seg (C19-CT)~\cite{Ma2021COVID-19-SegBenchmark} for abnormality classification and segmentation tasks, enabling a rigorous evaluation of cross-dataset generalization and robustness across acquisition domains.

\noindent\textbf{Experiment settings.} 
We pre-train the model on CT-RATE and evaluate it on 15 downstream datasets (Table~\ref{Talbe:dataset}), with external validation results reported in Sec.~\ref{Sec:external}.
CT volumetric scans are resampled to $1.5 \times 1.5 \times 3.0$ spacing, clipped to $[-1000,1000]$ HU, normalized to $[-1,1]$, and resized to $224 \times 224 \times 112$. The vision encoder is a 3D ViT-B~\cite{vit} initialized with MAE ImageNet-1K weights~\cite{He2021MAE}, paired with a 4-layer 3D ViT-B decoder (patch size $16 \times 16 \times 8$). The language branch uses a pre-trained CXR-BERT~\cite{boecking2022cxrbert} encoder and a 6-layer BERT decoder. 
Pre-training is conducted for 130 epochs using AdamW ($1.5 \times 10^{-4}$ learning rate, $0.05$ weight decay) on 4 NVIDIA A800 GPUs with batch size 14 per GPU for 15 hours, and masking probabilities $\alpha=\gamma=75\%$. For downstream tasks, the visual backbone is frozen except for segmentation, which is fine-tuned using a $96 \times 96 \times 48$ crop. All evaluations run on a single A800 or RTX 4090 GPU, with a total cost exceeding 800 GPU hours.

\begin{table*}[t]
\small
\centering
\setlength{\tabcolsep}{ 5.3pt}
\caption{Performance comparison on seven medical volumetric segmentation benchmarks spanning lesion and organ segmentation across CT and MRI modalities.
We report Dice similarity coefficient (DSC, \%). All methods are evaluated using a 3D ViT-B~\cite{vit} backbone under identical fine-tuning settings.
The best and second-best results are highlighted with \colorbox{bestresultcolor}{first} and \colorbox{subresultcolor}{second}, respectively.
${}^{\boldsymbol{\dagger}}$ indicates that the proposed method significantly outperforms the second-best approach under the same experimental setting, as assessed by a paired Wilcoxon signed-rank test ($p < 0.05$).}
\begin{tabular}{l|ccc|cc|cc|cc|c|c|c}
\toprule
\multirow{2}{*}{Method}  &\multicolumn{3}{c|}{LUNA16}  &\multicolumn{2}{c|}{MSD-Lung}  &\multicolumn{2}{c|}{COVID-19-20}  &\multicolumn{2}{c|}{SegThor} &\multicolumn{1}{c|}{FUMPE}  &\multicolumn{1}{c|}{BTCV}  &\multicolumn{1}{c}{ACDC}\\ 
&1\%  &10\%  &100\%  &10\%  &100\%  &10\%  &100\%  &10\%  &100\%  &100\%  &100\%  &100\%\\\midrule
Random init                     &$75.7_{5.5}$  &$89.1_{0.8}$  &$91.1_{0.5}$  &$25.6_{9.2}$  &$56.6_{4.8}$  &$32.7_{2.6}$  &$57.6_{2.9}$  &$60.8_{0.3}$  &$67.3_{0.5}$  &$40.1_{1.2}$  &$78.9_{0.3}$  &$82.6_{1.1}$\\\midrule
\multicolumn{3}{l}{\textit{\textcolor{gray}{3D Med-SSL}}} \\
MAE~\cite{He2021MAE}            &$79.0_{3.6}$  &$90.2_{0.5}$  &$92.0_{0.1}$  &$42.2_{4.8}$  &$65.5_{2.5}$  &$39.5_{1.7}$  &$64.0_{0.7}$  &$62.6_{0.3}$  &$71.5_{0.4}$  &$44.4_{1.4}$  &$79.7_{0.3}$  &$86.5_{0.3}$\\
PCRLv2~\cite{zhou2023pcrlv2}    &$79.7_{1.4}$  &$89.9_{0.4}$  &$92.0_{0.1}$  &$45.3_{2.4}$  &$66.9_{1.7}$  &$40.2_{1.5}$  &$63.9_{0.9}$  &$63.4_{0.3}$  &$71.9_{0.1}$  &$44.7_{0.1}$  &$80.1_{0.1}$  &$87.0_{0.2}$\\
HEMAE~\cite{tang2026hiendmae}   &\cellcolor{subresultcolor}$81.6_{0.9}$  &\cellcolor{subresultcolor}$90.6_{0.4}$  &$93.2_{0.1}$  &\cellcolor{subresultcolor}$53.1_{3.4}$  &\cellcolor{subresultcolor}$69.9_{0.3}$  &\cellcolor{subresultcolor}$42.1_{0.5}$  &\cellcolor{subresultcolor}$64.8_{0.3}$  &$64.0_{0.8}$  &\cellcolor{subresultcolor}$72.4_{0.2}$  &\cellcolor{subresultcolor}$45.4_{0.4}$  &\cellcolor{bestresultcolor}$81.7_{0.1}$  &$87.8_{0.1}$\\\midrule
\multicolumn{3}{l}{\textit{\textcolor{gray}{3D Med-VLP}}} \\
M3AE~\cite{chen2022m3ae}        &$79.4_{3.4}$  &$90.0_{0.4}$  &$92.2_{0.3}$  &$43.8_{3.2}$  &$67.1_{1.7}$  &$39.9_{1.5}$  &$64.5_{0.5}$  &$63.2_{0.4}$  &$71.7_{0.2}$  &$45.3_{0.1}$  &$80.0_{0.4}$  &$86.7_{0.2}$\\
MRM~\cite{zhou2023mrm}          &$79.7_{1.3}$  &$89.1_{0.6}$  &$92.6_{0.1}$  &$45.9_{4.6}$  &$68.5_{1.4}$  &$40.7_{1.6}$  &$63.9_{0.2}$  &$64.6_{0.4}$  &$71.5_{0.1}$  &$44.8_{0.1}$  &$80.1_{0.3}$  &$87.3_{1.1}$\\
CT-CLIP~\cite{hamamci2024ctclip}&$78.6_{2.0}$  &$89.7_{0.4}$  &$92.5_{0.1}$  &$45.6_{3.9}$  &$66.1_{1.8}$  &$38.6_{1.8}$  &$62.9_{0.7}$  &$64.1_{0.3}$  &$71.3_{0.2}$  &$43.8_{0.1}$  &$79.6_{0.2}$  &$86.3_{0.1}$\\
M3D~\cite{bai2024m3d}           &$79.2_{2.9}$  &$89.3_{0.5}$  &$92.3_{0.2}$  &$44.1_{5.4}$  &$64.5_{1.9}$  &$37.4_{0.9}$  &$62.5_{0.9}$  &$64.2_{0.2}$  &$71.4_{0.1}$  &$44.9_{0.1}$  &$79.1_{0.1}$  &$86.9_{0.2}$\\
fVLM~\cite{fvlm_iclr25}         &$80.0_{1.8}$  &$90.0_{0.8}$  &$92.5_{0.4}$  &$46.5_{2.0}$  &$69.4_{1.0}$  &$41.0_{1.1}$  &$64.4_{0.5}$  &$64.5_{0.2}$  &$72.0_{0.1}$  &$44.7_{0.0}$  &$79.9_{0.2}$  &$87.7_{0.2}$\\
HLIP~\cite{zhao2025hlip}        &$80.5_{1.6}$  &$89.7_{0.8}$  &$92.9_{0.1}$  &$46.2_{1.6}$  &$69.1_{0.6}$  &$39.9_{1.1}$  &$64.0_{0.2}$  &\cellcolor{subresultcolor}$64.7_{0.5}$  &$72.2_{0.1}$  &$44.4_{0.2}$  &$80.0_{0.9}$  &\cellcolor{subresultcolor}$88.1_{0.2}$\\
SimCroP~\cite{wang2025simcrop}  &$80.4_{2.3}$  &$90.3_{0.4}$  &\cellcolor{subresultcolor}$93.3_{0.1}$  &$50.1_{2.7}$  &$69.8_{0.8}$  &$41.9_{1.3}$  &$64.5_{0.5}$  &$63.2_{0.2}$  &$72.1_{0.2}$  &$44.9_{0.1}$  &\cellcolor{subresultcolor}$80.3_{0.1}$  &$87.6_{0.1}$\\
\rowcolor[gray]{1.0} \algname (Ours)  &\cellcolor{bestresultcolor}${83.7_{1.1}^{\boldsymbol{\dagger}}}$  &\cellcolor{bestresultcolor}${91.3_{0.2}^{\boldsymbol{\dagger}}}$  &\cellcolor{bestresultcolor}${93.4_{0.1}}$  &\cellcolor{bestresultcolor}${56.6_{2.4}^{\boldsymbol{\dagger}}}$  &\cellcolor{bestresultcolor}${70.9_{0.6}^{\boldsymbol{\dagger}}}$  &\cellcolor{bestresultcolor}${44.3_{1.6}^{\boldsymbol{\dagger}}}$  &\cellcolor{bestresultcolor}${65.1_{0.2}}$  &\cellcolor{bestresultcolor}${65.9_{0.1}^{\boldsymbol{\dagger}}}$  &\cellcolor{bestresultcolor}${72.8_{0.0}^{\boldsymbol{\dagger}}}$  &\cellcolor{bestresultcolor}${46.7_{0.1}^{\boldsymbol{\dagger}}}$  &\cellcolor{bestresultcolor}${81.7_{0.1}}$  &\cellcolor{bestresultcolor}${88.4_{0.1}}$\\\bottomrule
\end{tabular}
\vspace{-3mm}
\label{Table:Segmentation}
\end{table*}

\vspace{-4mm}
\subsection{Results on Downstream Tasks}
We evaluate ASAP on downstream tasks and comprehensively compare it with representative state-of-the-art pre-training methods~\cite{He2021MAE,chen2022m3ae,hamamci2024ctclip,zhou2023mrm,bai2024m3d,fvlm_iclr25,zhao2025hlip,wang2025simcrop,zhou2023pcrlv2,tang2026hiendmae,blankemeier2026merlin,cao2025vsd-boost} that provide publicly available implementations or pre-trained checkpoints. 
To ensure fair comparison of pre-training objectives, transformer-based methods are reproduced under a unified experimental protocol whenever possible. 
In particular, PCRLv2~\cite{zhou2023pcrlv2} and CT-CLIP~\cite{hamamci2024ctclip} are re-trained on the CT-RATE dataset using the same vision backbone and downstream evaluation protocol as ASAP, thereby minimizing the influence of architectural differences and isolating the effect of the pre-training strategy itself. 
Similarly, Hi-End-MAE (HEMAE)~\cite{tang2026hiendmae} is re-implemented and pre-trained on CT-RATE under the same data setting for controlled comparison. 
Methods originally proposed for 2D medical images~\cite{He2021MAE,chen2022m3ae,zhou2023mrm} are adapted to volumetric CT data following their official implementations and recommended training configurations. 
In addition to transformer-based approaches, we further include representative CNN-based methods, including Merlin~\cite{blankemeier2026merlin} and ViSD-Boost (ViSDB)~\cite{cao2025vsd-boost}, to provide a broader evaluation across different backbone paradigms and to contextualize the effectiveness of recent CT pre-training strategies beyond a single architectural family. 
All experiments are repeated five times with different random seeds, and the mean and standard deviation of each metric are reported.

\begin{table*}[t]
\small
\centering
\setlength{\tabcolsep}{ 5.5pt}
\caption{Performance comparison on the in-hospital mortality prediction task using the INSPECT dataset~\cite{huang2023inspect}. We report the area under AUC and concordance index (C-index), both in percentage. Except for methods with $^*$, all methods are evaluated using a 3D ViT-B~\cite{vit} backbone. The best and second-best results are highlighted with \colorbox{bestresultcolor}{first} and \colorbox{subresultcolor}{second}, respectively. ${}^{\boldsymbol{\dagger}}$ indicates that the proposed method significantly outperforms the second-best approach under the same experimental setting, based on a paired Wilcoxon signed-rank test ($p<0.05$).}
\begin{tabular}{l|ccc|ccc|ccc|ccc}
\toprule
\multirow{4}{*}{Method}  &\multicolumn{12}{c}{INSPECT}\\\cline{2-13}
&\multicolumn{9}{c|}{(AUC)}  &\multicolumn{3}{c}{\multirow{2}{*}{(C-Index)}}\\
&\multicolumn{3}{c|}{1\%}  &\multicolumn{3}{c|}{10\%}  &\multicolumn{3}{c|}{100\%}  &  &  &\\
&1M  &6M  &12M  &1M  &6M  &12M  &1M  &6M  &12M  &\multicolumn{1}{c}{1\%}  &\multicolumn{1}{c}{10\%}  &\multicolumn{1}{c}{100\%}\\\midrule
Random init                        &$64.7_{0.8}$  &$54.4_{0.6}$  &$56.1_{0.4}$  &$65.3_{0.6}$  &$63.9_{0.7}$  &$63.3_{0.3}$  &$71.1_{0.7}$  &$67.9_{0.2}$  &$66.1_{0.3}$  &$56.7_{0.5}$  &$62.9_{0.4}$  &$66.1_{0.2}$\\\midrule
\multicolumn{3}{l}{\textit{\textcolor{gray}{3D Med-SSL}}} \\
MAE~\cite{He2021MAE}               &$73.7_{0.8}$  &$68.7_{0.9}$  &$66.9_{0.8}$  &$78.1_{0.1}$  &$73.8_{0.2}$  &$71.4_{0.3}$  &$80.1_{0.1}$  &$76.3_{0.1}$  &$74.9_{0.2}$  &$67.9_{0.4}$  &$71.4_{0.1}$  &$74.2_{0.0}$\\
PCRLv2~\cite{zhou2023pcrlv2}       &$72.9_{0.5}$  &$69.5_{0.6}$  &$68.0_{0.2}$  &$78.9_{0.2}$  &$74.4_{0.2}$  &$72.0_{0.2}$  &$80.0_{0.1}$  &$77.0_{0.2}$  &$75.2_{0.2}$  &$68.6_{0.3}$  &$72.1_{0.1}$  &$74.9_{0.0}$\\
HEMAE~\cite{tang2026hiendmae}      &$73.5_{0.7}$  &$68.8_{0.7}$  &$66.8_{0.6}$  &$78.3_{0.4}$  &$74.0_{0.5}$  &$70.9_{0.4}$  &$79.4_{0.2}$  &$76.8_{0.1}$  &$74.7_{0.3}$  &$67.5_{0.3}$  &$71.0_{0.1}$  &$74.3_{0.0}$\\\midrule
\multicolumn{3}{l}{\textit{\textcolor{gray}{3D Med-VLP}}} \\
M3AE~\cite{chen2022m3ae}           &$73.8_{0.4}$  &$69.3_{0.4}$  &$70.2_{0.1}$  &$76.6_{0.8}$  &$74.5_{0.3}$  &$71.9_{0.2}$  &$79.4_{0.4}$  &$76.8_{0.4}$  &$75.6_{0.1}$  &$69.2_{0.0}$  &$72.0_{0.5}$  &$74.9_{0.0}$\\
MRM~\cite{zhou2023mrm}             &$74.7_{0.2}$  &$72.3_{0.3}$  &$70.5_{0.2}$  &$80.5_{0.1}$  &$77.7_{0.2}$  &$74.5_{0.1}$  &$81.9_{0.1}$  &$79.1_{0.2}$  &$76.9_{0.1}$  &$70.0_{0.2}$  &$74.9_{0.1}$  &$76.3_{0.0}$\\
CT-CLIP~\cite{hamamci2024ctclip}   &$75.1_{0.8}$  &$67.4_{0.5}$  &$67.2_{0.2}$  &$80.3_{0.2}$  &$76.8_{0.2}$  &$72.4_{0.6}$  &$82.2_{0.7}$  &$78.7_{0.7}$  &$76.5_{0.9}$  &$69.1_{0.3}$  &$74.3_{0.2}$  &$76.1_{0.8}$\\
M3D~\cite{bai2024m3d}              &$73.6_{0.4}$  &$68.1_{0.6}$  &$69.2_{0.4}$  &$78.3_{0.5}$  &$76.5_{0.5}$  &$75.4_{0.2}$  &$81.0_{0.2}$  &$78.3_{0.1}$  &$76.9_{0.1}$  &$69.7_{0.6}$  &$74.8_{0.1}$  &$76.3_{0.0}$\\
Merlin$^*$~\cite{blankemeier2026merlin}  &\cellcolor{subresultcolor}$78.7_{0.6}$  &\cellcolor{subresultcolor}$75.0_{0.4}$  &\cellcolor{subresultcolor}$73.2_{0.5}$  &$83.0_{0.1}$  &$79.8_{0.1}$  &$77.6_{0.1}$  &$83.8_{0.2}$  &$80.1_{0.3}$  &$77.7_{0.3}$  &\cellcolor{subresultcolor}$73.0_{0.4}$  &$77.3_{0.1}$  &$77.5_{0.3}$\\
fVLM~\cite{fvlm_iclr25}            &$78.5_{0.6}$  &$72.9_{0.3}$  &$70.8_{0.7}$  &$82.2_{0.1}$  &$78.8_{0.4}$  &$76.0_{0.7}$  &$84.0_{0.5}$  &$79.7_{0.2}$  &$77.9_{0.2}$  &$70.9_{0.5}$  &$75.5_{0.5}$  &$77.2_{0.3}$\\
ViSDB$^*$~\cite{cao2025vsd-boost}  &$77.9_{0.7}$  &$73.4_{0.5}$  &$71.7_{0.6}$  &$81.5_{0.2}$  &$77.4_{0.2}$  &$75.8_{0.1}$  &$82.9_{0.3}$  &$79.0_{0.2}$  &$77.4_{0.1}$  &$71.4_{0.5}$  &$75.1_{0.2}$  &$76.9_{0.2}$\\
HLIP~\cite{zhao2025hlip}           &$78.4_{0.4}$  &$73.7_{0.5}$  &$72.9_{0.5}$  &$82.4_{0.5}$  &$79.9_{0.3}$  &$77.6_{0.8}$  &$84.6_{0.4}$  &$80.9_{0.2}$  &$78.7_{0.3}$  &$72.6_{0.5}$  &$78.1_{0.4}$  &$78.2_{0.2}$\\
SimCroP~\cite{wang2025simcrop}     &$77.2_{0.6}$  &$73.5_{0.7}$  &$72.4_{0.5}$  &\cellcolor{subresultcolor}$83.1_{0.6}$  &\cellcolor{subresultcolor}$80.1_{0.7}$  &\cellcolor{subresultcolor}$78.0_{0.8}$  &\cellcolor{subresultcolor}$85.0_{0.2}$  &\cellcolor{subresultcolor}$81.3_{0.2}$  &\cellcolor{subresultcolor}$79.5_{0.2}$  &$72.2_{0.3}$  &\cellcolor{subresultcolor}$78.3_{0.2}$  &\cellcolor{subresultcolor}$79.0_{0.2}$\\
\rowcolor[gray]{1.0} \algname (Ours)&\cellcolor{bestresultcolor}$79.6_{0.3}^{\boldsymbol{\dagger}}$  &\cellcolor{bestresultcolor}$76.4_{0.2}^{\boldsymbol{\dagger}}$  &\cellcolor{bestresultcolor}$74.5_{0.2}^{\boldsymbol{\dagger}}$  &\cellcolor{bestresultcolor}$84.4_{0.1}^{\boldsymbol{\dagger}}$  &\cellcolor{bestresultcolor}$81.3_{0.0}^{\boldsymbol{\dagger}}$  &\cellcolor{bestresultcolor}$79.1_{0.1}^{\boldsymbol{\dagger}}$  &\cellcolor{bestresultcolor}$85.6_{0.0}^{\boldsymbol{\dagger}}$  &\cellcolor{bestresultcolor}$82.2_{0.1}^{\boldsymbol{\dagger}}$  &\cellcolor{bestresultcolor}$80.5_{0.2}^{\boldsymbol{\dagger}}$  &\cellcolor{bestresultcolor}$74.4_{0.2}^{\boldsymbol{\dagger}}$  &\cellcolor{bestresultcolor}$78.6_{0.1}^{\boldsymbol{\dagger}}$  &\cellcolor{bestresultcolor}$80.0_{0.1}^{\boldsymbol{\dagger}}$\\\bottomrule
\end{tabular}
\vspace{-5mm}
\label{Table:prognose}
\end{table*}

\begin{figure}[t]
    \centering
    \includegraphics[width=\linewidth]{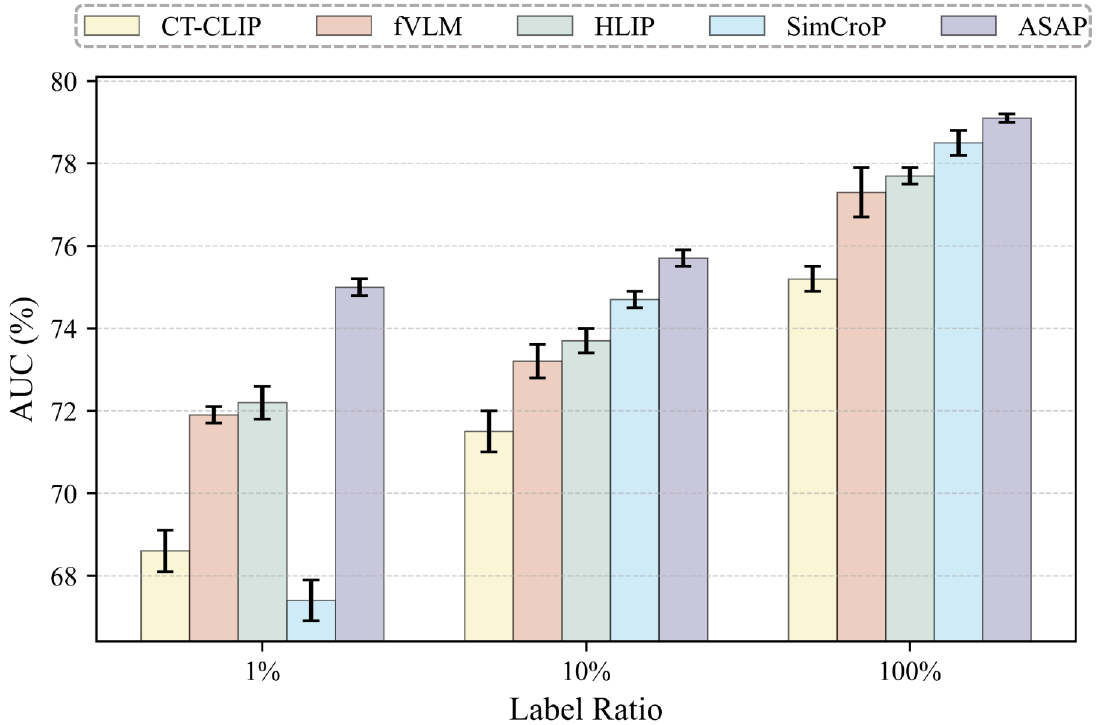}
    \caption{Quantitative comparison on the severe disease prediction task on the STOIC-2021 dataset~\cite{revel2021stoic}. The AUC (\%) is reported under varying label ratios, reflecting different levels of annotation availability. Error bars represent the standard deviation over five independent runs.}
    \vspace{-6mm}
    \label{fig:stoic}
\end{figure}

\begin{table}[t]
\small
\centering
\setlength{\tabcolsep}{ 2.4pt}
\caption{Performance comparison on medical volumetric report generation task using CTRG~\cite{tang2024ctrg} and CT-Rate~\cite{hamamci2024ctclip} benchmark. We report BLEU scores (BL) and BERTScore (B-S), all in percentage. All methods are evaluated with a 3D ViT-B~\cite{vit} backbone. The best and second-best results are highlighted with \colorbox{bestresultcolor}{first} and \colorbox{subresultcolor}{second}, respectively. ${}^{\boldsymbol{\dagger}}$ indicates that the proposed method significantly outperforms the second-best approach under the same experimental setting, based on a paired Wilcoxon signed-rank test ($p<0.05$).}
\begin{tabular}{l|l|ccccc}
\toprule
\multirow{1}{*}{Dataset}  &\multirow{1}{*}{Method}  &\multicolumn{1}{c}{BL-1}  &\multicolumn{1}{c}{BL-2}  &\multicolumn{1}{c}{BL-3}  &\multicolumn{1}{c}{BL-4}  &\multicolumn{1}{c}{B-S}\\
\midrule
\multirow{14}{*}{CTRG}  &Random init  &$55.5_{0.4}$  &$45.1_{0.5}$  &$38.9_{0.4}$  &$35.1_{0.4}$  &$75.4_{0.1}$\\\cline{2-7}
&\multicolumn{5}{l}{\textit{\textcolor{gray}{3D Med-SSL}}} \\
&MAE~\cite{He2021MAE}  &$56.4_{0.4}$  &$45.5_{0.1}$  &$39.2_{0.4}$  &$35.2_{0.1}$  &$75.8_{0.2}$\\
&PCRLv2~\cite{zhou2023pcrlv2}  &$59.5_{0.3}$  &$48.9_{0.5}$  &$41.8_{0.4}$  &$36.4_{0.2}$  &$77.6_{0.2}$\\
&HEMAE~\cite{tang2026hiendmae}  &$59.2_{0.3}$  &$47.8_{0.1}$  &$40.9_{0.2}$  &$36.1_{0.2}$  &$76.1_{0.1}$\\\cline{2-7}
&\multicolumn{5}{l}{\textit{\textcolor{gray}{3D Med-VLP}}} \\
&M3AE~\cite{chen2022m3ae}  &$58.3_{0.2}$  &$47.3_{0.3}$  &$40.7_{0.1}$  &$36.3_{0.1}$  &$76.3_{0.4}$\\
&MRM~\cite{zhou2023mrm}  &$61.0_{0.5}$  &$49.8_{0.5}$  &$42.3_{0.3}$  &$37.2_{0.4}$  &$78.2_{0.2}$\\
&CT-CLIP~\cite{hamamci2024ctclip}  &$56.3_{0.1}$  &$45.5_{0.3}$  &$39.2_{0.4}$  &$35.2_{0.1}$  &$76.4_{0.3}$\\
&M3D~\cite{bai2024m3d}  &$58.5_{0.1}$  &$47.7_{0.1}$  &$40.6_{0.5}$  &$35.6_{0.5}$  &$77.6_{0.1}$\\
&fVLM~\cite{fvlm_iclr25}  &$60.6_{0.1}$  &$49.6_{0.4}$  &$42.4_{0.2}$  &$37.3_{0.1}$  &$78.9_{0.3}$\\
&HLIP~\cite{zhao2025hlip}  &\cellcolor{subresultcolor}$61.1_{0.4}$  &\cellcolor{subresultcolor}$49.9_{0.1}$  &\cellcolor{subresultcolor}$42.6_{0.3}$  &\cellcolor{subresultcolor}$37.6_{0.1}$  &\cellcolor{subresultcolor}$79.2_{0.2}$\\
&SimCroP~\cite{wang2025simcrop}  &$59.7_{0.5}$  &$48.4_{0.5}$  &$40.9_{0.2}$  &$35.7_{0.2}$  &$78.1_{0.2}$\\
&\algname (Ours)  &\cellcolor{bestresultcolor}$62.8_{0.2}^{\boldsymbol{\dagger}}$  &\cellcolor{bestresultcolor}$51.3_{0.2}^{\boldsymbol{\dagger}}$  &\cellcolor{bestresultcolor}$43.7_{0.1}^{\boldsymbol{\dagger}}$  &\cellcolor{bestresultcolor}$38.3_{0.3}^{\boldsymbol{\dagger}}$  &\cellcolor{bestresultcolor}$79.5_{0.2}$\\\midrule
\multirow{14}{*}{CT-Rate}  &Random init  &$56.5_{0.1}$  &$44.1_{0.1}$  &$37.1_{0.2}$  &$32.2_{0.1}$  &$74.5_{0.4}$\\\cline{2-7}
&\multicolumn{5}{l}{\textit{\textcolor{gray}{3D Med-SSL}}} \\
&MAE~\cite{He2021MAE}  &$58.4_{0.2}$  &$45.9_{0.3}$  &$38.2_{0.2}$  &$33.0_{0.4}$  &$75.4_{0.1}$\\
&PCRLv2~\cite{zhou2023pcrlv2}  &$59.0_{0.1}$  &$46.5_{0.1}$  &$38.4_{0.2}$  &$33.7_{0.4}$  &$75.6_{0.3}$\\
&HEMAE~\cite{tang2026hiendmae}  &$58.3_{0.2}$  &$45.9_{0.1}$  &$38.0_{0.1}$  &$32.8_{0.2}$  &$75.2_{0.4}$\\\cline{2-7}
&\multicolumn{5}{l}{\textit{\textcolor{gray}{3D Med-VLP}}} \\
&M3AE~\cite{chen2022m3ae}  &$58.3_{0.1}$  &$45.9_{0.3}$  &$38.6_{0.3}$  &$33.5_{0.1}$  &$75.5_{0.3}$\\
&MRM~\cite{zhou2023mrm}  &$58.2_{0.2}$  &$45.9_{0.1}$  &$38.8_{0.1}$  &$33.8_{0.2}$  &\cellcolor{subresultcolor}$75.8_{0.1}$\\
&CT-CLIP~\cite{hamamci2024ctclip}  &$58.1_{0.2}$  &$45.7_{0.1}$  &$38.5_{0.2}$  &$33.9_{0.2}$  &$75.4_{0.3}$\\
&M3D~\cite{bai2024m3d}  &$57.8_{0.1}$  &$46.3_{0.3}$  &$38.9_{0.1}$  &$33.8_{0.3}$  &$75.5_{0.2}$\\
&fVLM~\cite{fvlm_iclr25}  &$58.6_{0.2}$  &$46.2_{0.3}$  &$38.9_{0.5}$  &$34.1_{0.1}$  &$75.5_{0.5}$\\
&HLIP~\cite{zhao2025hlip}  &\cellcolor{subresultcolor}$59.8_{0.4}$  &\cellcolor{subresultcolor}$46.9_{0.2}$  &\cellcolor{subresultcolor}$39.0_{0.2}$  &\cellcolor{subresultcolor}$34.2_{0.1}$  &\cellcolor{bestresultcolor}$75.9_{0.3}$\\
&SimCroP~\cite{wang2025simcrop}  &$58.2_{0.4}$  &$45.9_{0.2}$  &$38.5_{0.3}$  &$33.9_{0.2}$  &$75.4_{0.2}$\\
&\algname (Ours)  &\cellcolor{bestresultcolor}$60.0_{0.1}$  &\cellcolor{bestresultcolor}$47.2_{0.3}$  &\cellcolor{bestresultcolor}$39.2_{0.3}$  &\cellcolor{bestresultcolor}$34.4_{0.2}$  &\cellcolor{subresultcolor}$75.8_{0.2}$\\
\bottomrule
\end{tabular}
\label{Talbe:report_gen}
\vspace{-5mm}
\end{table}

\begin{table}[t]
\small
\centering
\setlength{\tabcolsep}{ 2.6pt}
\caption{Performance comparison on vocabulary classification and volume–text retrieval tasks using the CT-RATE~\cite{hamamci2024ctclip} and AHPH-10K benchmarks. We report AUC for vocabulary classification and Recall@50 for retrieval, both in percentage. All methods are evaluated with a 3D ViT-B~\cite{vit} backbone. The best and second-best results are highlighted with \colorbox{bestresultcolor}{first} and \colorbox{subresultcolor}{second}, respectively. ${}^{\boldsymbol{\dagger}}$ indicates that the proposed method significantly outperforms the second-best approach under the same experimental setting, based on a paired Wilcoxon signed-rank test ($p<0.05$).}
\begin{tabular}{l|ccc|ccc}
\toprule
\multirow{3}{*}{Method}  &\multicolumn{3}{c|}{CT-Rate}  &\multicolumn{3}{c}{AHPH-10K}\\
&\multicolumn{1}{c}{Voc}  &\multicolumn{2}{c|}{Ret (R@50)}  &\multicolumn{1}{c}{Voc}  &\multicolumn{2}{c}{Ret (R@50)}\\
&(AUC)  &\multicolumn{1}{c}{V2R}  &\multicolumn{1}{c|}{R2V}  &(AUC)  &\multicolumn{1}{c}{V2R}  &\multicolumn{1}{c}{R2V}\\
\midrule
Random init                            &$60.6_{0.4}$  &$7.7_{0.2}$  &$8.4_{0.3}$  &$51.6_{0.3}$  &$4.7_{0.2}$  &$4.3_{0.3}$\\\midrule
\multicolumn{3}{l}{\textit{\textcolor{gray}{3D Med-SSL}}} \\
MAE~\cite{He2021MAE}                   &$72.7_{0.2}$  &$17.6_{0.1}$  &$17.2_{0.2}$  &$58.4_{0.5}$  &$8.4_{0.3}$  &$8.6_{0.2}$\\
PCRLv2~\cite{zhou2023pcrlv2}           &$73.2_{0.1}$  &$18.9_{0.2}$  &$18.8_{0.1}$  &$60.2_{0.4}$  &$9.0_{0.1}$  &$8.7_{0.1}$\\
HEMAE~\cite{tang2026hiendmae}          &$70.8_{0.3}$  &$14.3_{0.1}$  &$14.0_{0.2}$  &$56.5_{0.3}$  &$7.2_{0.2}$  &$7.0_{0.3}$\\\midrule
\multicolumn{3}{l}{\textit{\textcolor{gray}{3D Med-VLP}}} \\
M3AE~\cite{chen2022m3ae}               &$73.5_{0.3}$  &$19.6_{0.2}$  &$19.0_{0.3}$  &$59.9_{0.5}$  &$8.9_{0.3}$  &$8.7_{0.2}$\\
MRM~\cite{zhou2023mrm}                 &$73.9_{0.1}$  &$20.5_{0.2}$  &$19.6_{0.1}$  &$62.9_{0.4}$  &$8.9_{0.1}$  &$9.0_{0.3}$\\
CT-CLIP~\cite{hamamci2024ctclip}       &$75.0_{0.1}$  &$16.5_{0.1}$  &$17.1_{0.2}$  &$63.9_{0.3}$  &$8.3_{0.2}$  &$7.1_{0.1}$\\
M3D~\cite{bai2024m3d}                  &$74.5_{0.4}$  &$15.6_{0.1}$  &$15.0_{0.1}$  &$63.7_{0.2}$  &$7.9_{0.4}$  &$8.0_{0.3}$\\
fVLM~\cite{fvlm_iclr25}                &$75.3_{0.1}$  &$27.9_{0.2}$  &$27.2_{0.1}$  &$64.3_{0.6}$  &$9.4_{0.1}$  &$9.2_{0.1}$\\
HLIP~\cite{zhao2025hlip}               &$76.6_{0.2}$  &$30.3_{0.1}$  &$29.9_{0.1}$  &\cellcolor{subresultcolor}$65.5_{0.5}$  &\cellcolor{subresultcolor}$11.7_{0.2}$  &\cellcolor{subresultcolor}$11.4_{0.3}$\\
SimCroP~\cite{wang2025simcrop}         &\cellcolor{subresultcolor}$77.2_{0.2}$  &\cellcolor{subresultcolor}$35.1_{0.3}$  &\cellcolor{subresultcolor}$34.0_{0.1}$  &$64.9_{0.8}$  &$11.2_{0.1}$  &$10.9_{0.1}$\\
\algname (Ours)   &\cellcolor{bestresultcolor}{$78.4_{0.1}^{\boldsymbol{\dagger}}$}  &\cellcolor{bestresultcolor}{$37.0_{0.1}^{\boldsymbol{\dagger}}$}  &\cellcolor{bestresultcolor}{$36.5_{0.1}^{\boldsymbol{\dagger}}$}  &\cellcolor{bestresultcolor}{$69.2_{0.4}^{\boldsymbol{\dagger}}$}  &\cellcolor{bestresultcolor}{$13.6_{0.2}^{\boldsymbol{\dagger}}$}  &\cellcolor{bestresultcolor}{$12.9_{0.2}^{\boldsymbol{\dagger}}$}\\
\bottomrule
\end{tabular}
\vspace{-5mm}
\label{Talbe:vocabfine_retrieval}
\end{table}

\vspace{-2mm}
\subsubsection{Medical Volumetric Abnormality Classification}
As summarized in Table~\ref{Table:LinearProbe}, \algname achieves the best overall performance on volumetric abnormality classification across six benchmarks and multiple supervision regimes, outperforming representative Med-SSL and Med-VLP methods~\cite{He2021MAE,chen2022m3ae,hamamci2024ctclip,zhou2023mrm,bai2024m3d,fvlm_iclr25,zhao2025hlip,wang2025simcrop,zhou2023pcrlv2,tang2026hiendmae,blankemeier2026merlin,cao2025vsd-boost}.
The advantage is more pronounced under extremely limited supervision. With only 1\% labeled data, \algname improves the AUC over the strongest competing baseline by 2.2\% on STOIC-2021~\cite{revel2021stoic} and 2.7\% on Rad-ChestCT~\cite{draelos2021radchestct}.
The improvements are statistically significant ($p<0.05$) across all supervision regimes on five of the six benchmarks, demonstrating consistent gains across repeated runs.
Compared with our prior variant using a similar backbone, \algname further provides stable improvements, highlighting the benefit of the proposed pre-training strategy.
\vspace{-2mm}
\subsubsection{Medical Volumetric Segmentation}
Table~\ref{Table:Segmentation} summarizes volumetric segmentation results on seven benchmarks covering thoracic CT, abdominal CT, and cross-modality CT-to-MRI settings.
We adopt UNETR~\cite{hatamizadeh2022unetr} as the segmentation framework, where the 3D ViT-B encoder is initialized using different pre-training methods.
\algname achieves the best overall performance across most datasets and supervision regimes. The advantage is particularly evident in low-data settings. For instance, on MSD-Lung~\cite{antonelli2022msd}, \algname improves the Dice score by 3.5\% over the strongest competing baseline with only 10\% labeled data.
Despite being pre-trained solely on chest CT data, \algname generalizes effectively to anatomically distinct abdominal CT datasets and remains competitive on ACDC~\cite{bernard2018acdc}, which involves substantial anatomical and modality shifts.
Furthermore, the improvements are statistically significant across all supervision regimes on five of the seven benchmarks ($p<0.05$), indicating consistent gains across repeated runs.
These results demonstrate the strong transferability of the proposed volumetric representations across datasets, organs, and imaging modalities.

\vspace{-3mm}
\subsubsection{Medical Volumetric Prognosis Prediction}
Beyond diagnostic tasks, we further evaluate the learned representations on prognosis prediction, which requires sensitivity to subtle and spatially sparse patterns associated with disease progression. 
Experiments are conducted on INSPECT~\cite{huang2023inspect} for in-hospital mortality prediction and STOIC-2021~\cite{revel2021stoic} for COVID-19 severity assessment, following the official evaluation protocols. 
As reported in Table~\ref{Table:prognose} and Fig.~\ref{fig:stoic}, \algname consistently achieves the best overall performance across evaluated settings, with particularly pronounced gains under limited supervision. 
With only 1\% labeled data, \algname improves the C-Index on INSPECT by 1.8\% over the strongest competing baseline, indicating stronger transferability for prognostic representation learning. 
Notably, Merlin~\cite{blankemeier2026merlin}, a method specifically designed for prognosis-oriented pre-training, achieves competitive performance in the extremely low-data regime but shows comparatively smaller improvements as the amount of labeled data increases. 
In contrast, \algname maintains consistent gains across supervision regimes, suggesting better scalability and representation generalization for downstream prognosis prediction. 
Moreover, \algname consistently outperforms our prior variant SimCroP~\cite{wang2025simcrop} with statistically significant margins ($p<0.05$), further supporting the effectiveness of semantically-adaptive alignment and anatomy-aware knowledge injection for modeling fine-grained pathological progression patterns.

\vspace{-3mm}
\subsubsection{Medical Volumetric Report Generation} 
Table~\ref{Talbe:report_gen} presents report generation results on CTRG-Chest~\cite{tang2024ctrg} and CT-Rate~\cite{hamamci2024ctclip}, evaluated using BLEU and BERTScore~\cite{zhang2020bertscore}. 
Following prior work~\cite{wu2025vocolarge,ni2024mg3d}, we replace only the vision encoder while keeping the language model fixed as in M2KT~\cite{yang2023m2kt}.
\algname achieves the best overall performance across both benchmarks, with statistically significant gains ($p<0.05$) on CTRG-Chest.
The results demonstrate the effectiveness of the proposed pre-training strategy for volumetric report generation.

\vspace{-2.5mm}
\subsubsection{Vocabulary Classification and Volume--text Retrieval}
Table~\ref{Talbe:vocabfine_retrieval} reports vocabulary classification and volume--text retrieval results on CT-Rate~\cite{hamamci2024ctclip} and AHPH-10K.
Following prior protocols~\cite{wu2025vocolarge,ni2024mg3d}, we replace only the vision encoder while keeping the language model and training strategy identical to CT-CLIP~\cite{hamamci2024ctclip}.
Across both datasets, \algname achieves the best overall performance on classification (AUC) and retrieval (Recall@50) metrics.
On AHPH-10K, \algname further obtains statistically significant improvements ($p<0.05$), including a 3.7\% AUC gain and a 1.7\% average Recall@50 improvement over the strongest competing baseline.
These results suggest improved transferability of the learned volumetric representations under cross-dataset settings.

\begin{figure*}[t]
    \centering
    \includegraphics[width=0.9\linewidth]{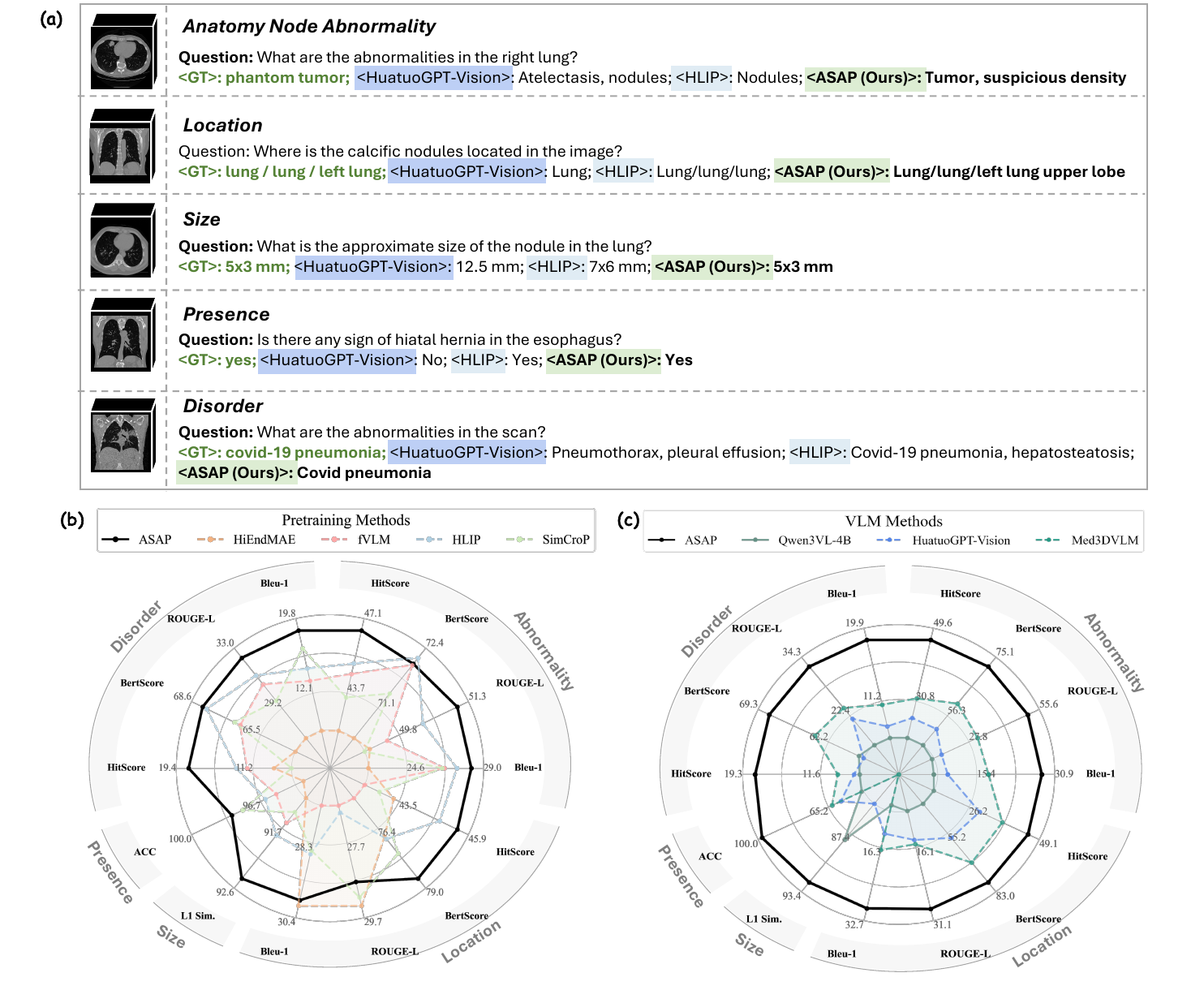}
    \caption{Examples and quantitative comparisons for medical volumetric visual question answering on the RadGenome-Chest CT dataset~\cite{zhang2025radgenomect}. (a) Qualitative results across five sub-tasks, including abnormality detection, localization, size estimation, presence prediction, and disorder prediction. 
    (b) Comparison among different pre-training strategies under the same LLM-based evaluation protocol. 
    (c) Comparison between \algname and existing vision--language models (VLMs), including HuatuoGPT-Vision~\cite{chen2024huatuogptvision} and Med3DVLM~\cite{yu2025med3dvlm}, under identical settings. 
    Radar plots report performance across multiple metrics (e.g., BLEU, ROUGE-L, BERTScore~\cite{zhang2020bertscore}, HitScore, and task-specific accuracy).}
    \vspace{-5mm}
    \label{fig:vqa}
\end{figure*}

\begin{figure*}[t]
    \centering
    \includegraphics[width=\linewidth]{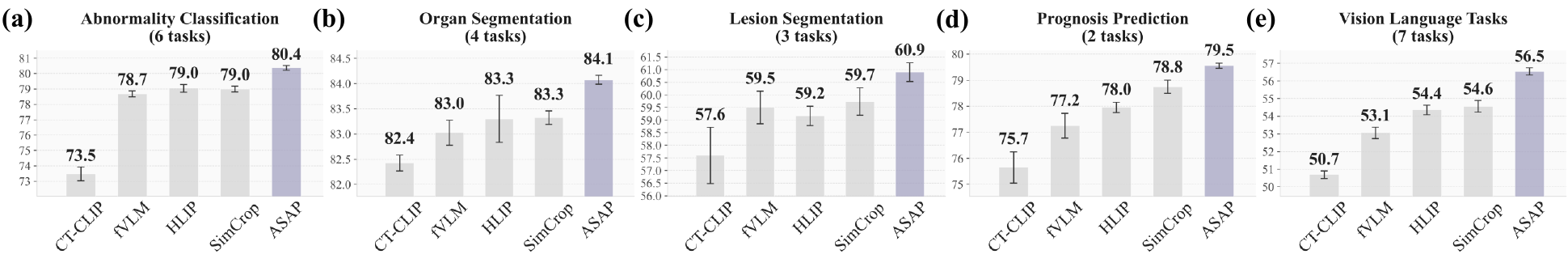}
    \caption{Overall comparison on our benchmark. We evaluate different pre-training methods across five task groups, namely abnormality classification, organ segmentation, lesion segmentation, prognosis prediction, and vision-language tasks. For each group, the reported results correspond to the average performance over all tasks within that group. Error bars represent the standard deviation across multiple runs.}
    \label{fig:overall_bar}
    \vspace{-3mm}
\end{figure*}

\begin{table*}[t]
\centering
\setlength{\tabcolsep}{ 7pt}
\caption{Cross-dataset external validation results under a transfer setting. The benchmark covers both abnormality classification (AUC) and volumetric segmentation (Dice) tasks across five source$\rightarrow$target pairs. The best and second-best results are highlighted with \colorbox{bestresultcolor}{first} and \colorbox{subresultcolor}{second}, respectively.}
\label{Table:external}
\begin{tabular}{l|ccc|cc|cc|ccc|cc}
\toprule
\multirow{4}{*}{Methods}  &\multicolumn{12}{c}{\textbf{Source$\rightarrow$Target}}\\\cline{2-13}
&\multicolumn{3}{c|}{CT-Rate$\rightarrow$AHPH-10K}  &\multicolumn{2}{c|}{INSPECT$\rightarrow$RSPECT}  &\multicolumn{2}{c|}{MSD-Lung$\rightarrow$RIDER}  &\multicolumn{3}{c|}{LUNA16$\rightarrow$C19-CT}  &\multicolumn{2}{c}{COVID-19-20$\rightarrow$C19-CT}\\ 
&\multicolumn{3}{c|}{(AUC)}  &\multicolumn{2}{c|}{(AUC)}  &\multicolumn{2}{c|}{(Dice)}  &\multicolumn{3}{c|}{(Dice)}  &\multicolumn{2}{c}{(Dice)}\\
&1\%  &10\%  &100\%  &10\%  &100\%  &10\%  &100\%  &1\%  &10\%  &100\%  &10\%  &100\%\\\midrule 
HEMAE~\cite{tang2026hiendmae}        &62.2  &63.0  &65.1  &59.9  &61.8  &19.2  &\cellcolor{subresultcolor}{40.5}  &13.9  &73.8  &88.8  &64.4  &74.4\\\midrule
CT-CLIP~\cite{hamamci2024ctclip}     &61.9  &62.9  &65.5  &58.3  &61.2  &18.1  &31.7  &10.4  &65.9  &88.9  &64.2  &73.9\\
fVLM~\cite{fvlm_iclr25}              &65.3  &68.0  &68.4  &61.6  &62.2  &\cellcolor{subresultcolor}{21.2}  &36.8  &20.1  &66.9  &\cellcolor{subresultcolor}{90.9}  &60.7  &74.5\\
HLIP~\cite{zhao2025hlip}             &65.5  &\cellcolor{subresultcolor}{68.4}  &\cellcolor{subresultcolor}{69.1}  &61.6  &62.5  &19.8  &38.8  &16.1  &\cellcolor{subresultcolor}{81.4}  &89.9  &\cellcolor{subresultcolor}{64.9}  &72.8\\
SimCroP~\cite{wang2025simcrop}       &\cellcolor{subresultcolor}{67.4}  &\cellcolor{subresultcolor}{68.4}  &68.9  &\cellcolor{subresultcolor}{61.8}  &\cellcolor{subresultcolor}{62.8}  &20.9  &38.7  &\cellcolor{subresultcolor}{20.3}  &72.3  &87.7  &61.2  &\cellcolor{subresultcolor}{75.0}\\\midrule
\algname (Ours)                      &\cellcolor{bestresultcolor}{$68.2$}  &\cellcolor{bestresultcolor}{69.5}  &\cellcolor{bestresultcolor}{69.8}  &\cellcolor{bestresultcolor}{62.0}  &\cellcolor{bestresultcolor}{63.1}  &\cellcolor{bestresultcolor}{21.8}  &\cellcolor{bestresultcolor}{42.7}  &\cellcolor{bestresultcolor}{21.3}  &\cellcolor{bestresultcolor}{92.4}  &\cellcolor{bestresultcolor}{92.6}  &\cellcolor{bestresultcolor}{67.3}  &\cellcolor{bestresultcolor}{75.1}\\\bottomrule
\end{tabular}
\vspace{-5mm}
\end{table*}

\vspace{-4mm}
\subsubsection{Medical Volumetric Visual Question Answering}
Fig.~\ref{fig:vqa} presents the open-ended visual question answering (VQA) results on the RadGenome-Chest CT dataset~\cite{zhang2025radgenomect} across five official sub-tasks. 
This experiment aims to assess whether the proposed pre-training method facilitates effective integration with large language models (LLMs) for downstream multi-modal reasoning. 
To this end, we connect the pre-trained visual encoder to a general-purpose LLM (Qwen3-4B-Instruct~\cite{yang2025qwen3}) via a Q-former and a lightweight projection module, without modifying the LLM architecture or performing task-specific architectural redesign. 
Under this setting, \algname achieves competitive performance using only 2,000 training samples\footnote{The official subset of RadGenome-Chest CT released in \url{https://github.com/xiaoman-zhang/RadGenome-ChestCT}.} for supervised fine-tuning (SFT), surpassing the performance of specialized medical VLMs such as HuatuoGPT-Vision~\cite{chen2024huatuogptvision} and Med3DVLM~\cite{yu2025med3dvlm} under comparable evaluation protocols. 
These results suggest that the representations learned by \algname are readily adaptable to LLM-based multi-modal frameworks and can support effective cross-modal reasoning with minimal supervision. 
In particular, the strong performance in the low-data regime indicates improved data efficiency and more precise cross-modal grounding, which are critical for open-ended clinical reasoning tasks. 
Overall, this experiment demonstrates that the proposed pre-training strategy not only enhances standalone representation quality but also enables effective and modular integration with external LLMs for downstream medical applications.
\vspace{-2mm}
\subsubsection{Discussion}
\label{sec:disscusion}
\textbf{Overall improvements.} 
As shown in Fig.~\ref{fig:overall_bar}, \algname consistently outperforms representative Med-VLP methods~\cite{hamamci2024ctclip,fvlm_iclr25,zhao2025hlip,wang2025simcrop} under a unified evaluation protocol. 
Compared with SimCroP~\cite{wang2025simcrop}, the conference version, \algname achieves consistent gains across classification, segmentation, prognosis prediction, and vision--language tasks. 
Notably, the improvements are statistically significant across task categories (paired Wilcoxon signed-rank test, $p=6.72\times10^{-7}$), suggesting that the proposed framework improves general-purpose volumetric representation learning rather than benefiting only isolated downstream settings. 
These results indicate that anatomy-aware structural modeling and semantically-adaptive alignment provide complementary advantages for learning transferable volumetric representations across heterogeneous clinical tasks.

\textbf{Label efficiency.} 
Across 12 downstream tasks with varying supervision ratios, \algname consistently achieves strong performance under limited labeled data. 
In several classification, segmentation, and prognosis prediction tasks, using only 1\% or 10\% labeled data yields performance comparable to or exceeding prior methods trained with full supervision. 
Similar trends are observed in multimodal reasoning tasks on RadGenome-Chest CT~\cite{zhang2025radgenomect}, where \algname maintains competitive VQA performance under reduced training data. 
These observations suggest that the proposed pre-training strategy learns more transferable and data-efficient volumetric representations, particularly in low-supervision settings where representation quality becomes more critical.


\vspace{-3mm}
\textbf{Failure cases.} 
Although \algname achieves consistent improvements across most downstream tasks, the gains are less pronounced in several settings involving substantial distribution shifts or complex vision--language generation objectives. 
For example, performance improvements on pulmonary embolism classification using CTPA data~\cite{huang2023inspect} and segmentation on abdominal CT~\cite{landman2015btcv} or cardiac MRI benchmarks~\cite{bernard2018acdc} remain limited, likely due to differences in imaging protocols, anatomical distributions, and imaging modalities relative to the chest CT data used during pre-training. 
In addition, improvements on report generation and certain VQA metrics are comparatively modest. 
This suggests that representation learning alone may be insufficient for complex multi-modal generation and understanding tasks, where language modeling capacity, decoding strategies, and fine-grained reasoning also play important roles. 
Moreover, different pre-training objectives may emphasize different aspects of multi-modal alignment and language generation, leading to metric-specific advantages across VQA subtasks. 
Overall, these findings highlight remaining challenges in cross-domain generalization and the joint optimization of visual grounding, understanding and language generation for medical vision--language pre-training.

\vspace{-3mm}
\subsection{External Validation}
\label{Sec:external}
We further evaluate the cross-dataset transferability of different pre-training methods under an external validation setting. 
Specifically, models are first fine-tuned on five source datasets and then directly transferred to four unseen target datasets. The evaluation includes both abnormality classification (AUC) and volumetric segmentation (Dice) across multiple source$\rightarrow$target pairs. 
As shown in Table~\ref{Table:external}, \algname consistently achieves the best overall performance across different transfer settings. The improvements are particularly evident in low-data regimes, where transferability is more critical. 
Overall, the results suggest that the proposed pre-training strategy learns more transferable volumetric representations under cross-dataset distribution shifts, benefiting both classification and segmentation tasks without target-domain adaptation.

\begin{table*}[h] 
\centering
\setlength{\tabcolsep}{ 6pt}
\caption{Ablation study on each design component in our framework on abnormality classification (RadChestCT), lesion segmentation (MSD-Lung~\cite{antonelli2022msd}), organ segmentation (SegThor~\cite{lambert2020segthor}), prognosis prediction (INSPECT~\cite{huang2023inspect}), report generation (CTRG~\cite{tang2024ctrg}), and vocabulary classification \& volume-report retrieval (AHPH-10K). ``\textit{AKI}" refers to ``\textit{Anatomy-aware Knowledge Injection}", ``\textit{SSA}" denotes ``\textit{Semantically-adaptive Selective Alignment}", and ``\textit{Fusion}" stands for ``\text{Semantically-adaptive Fusion}", which is consisted of ``\textit{Semantic-augmented Localized Fusion}" (``\textit{SLF}") and ``\textit{Global Fusion}" (``\textit{GF}"). \CheckmarkBold denotes the component is included. The best results are \textbf{bolded}.}
\label{Table:ablation}
\begin{tabular}{cccc|cc|cc|cc|ccc|cc|cc}
\toprule
\multirow{3}{*}{AKI}  &\multirow{3}{*}{SSA}  &\multicolumn{2}{c|}{\multirow{2}{*}{Fusion}}  &\multicolumn{2}{c|}{RadChestCT}  &\multicolumn{2}{c|}{MSD-Lung}  &\multicolumn{2}{c|}{SegThor}  &\multicolumn{3}{c|}{INSPECT}  &\multicolumn{2}{c|}{CTRG}  &\multicolumn{2}{c}{AHPH-10K}\\ 
&  &  &  &\multicolumn{2}{c|}{(AUC)}  &\multicolumn{2}{c|}{(Dice)}  &\multicolumn{2}{c|}{(Dice)}  &\multicolumn{3}{c|}{(C-Index)}  &\multicolumn{1}{c}{(BL-4)}  &\multicolumn{1}{c|}{(B-S)}  &\multicolumn{1}{c}{(AUC)}  &\multicolumn{1}{c}{(R@50)}\\\cline{3-4}
&  &\multicolumn{1}{c}{SLF}  &\multicolumn{1}{c|}{GF}  &10\%  &100\%  &10\%  &100\%  &10\%  &100\%  &1\%  &10\%  &100\%  &\multicolumn{2}{c|}{100\%}  &\multicolumn{2}{c}{100\%}\\\midrule 
  &  &  &\CheckmarkBold                                   &$71.6$  &$73.8$  &$44.2$  &$66.4$  &$63.9$  &$71.7$  &$70.2$  &$74.4$  &$78.4$  &$36.6$  &$77.8$  &$62.5$  &$9.0$\\
\CheckmarkBold  &  &  &\CheckmarkBold                     &$73.1$  &$75.5$  &$47.5$  &$67.9$  &$64.7$  &$72.3$  &$71.3$  &$75.3$  &$78.7$  &$37.0$  &$78.4$  &$65.9$  &$11.6$\\\midrule
  &\CheckmarkBold  &  &\CheckmarkBold                     &$74.1$  &$76.5$  &$48.9$  &$68.7$  &$64.4$  &$72.0$  &$72.7$  &$77.3$  &$79.3$  &$37.3$  &$78.5$  &$66.6$  &$11.8$\\
  &\CheckmarkBold  &\CheckmarkBold  &                     &$73.6$  &$76.1$  &$50.0$  &$69.1$  &$64.9$  &$72.2$  &$72.0$  &$76.8$  &$78.9$  &$37.6$  &$78.9$  &$66.4$  &$12.3$\\
  &\CheckmarkBold  &\CheckmarkBold  &\CheckmarkBold       &$74.7$  &$76.8$  &$50.7$  &$69.8$  &$64.7$  &$72.1$  &$72.9$  &$77.5$  &$79.5$  &$38.0$  &$79.1$  &$66.8$  &$12.1$\\\midrule
\CheckmarkBold  &\CheckmarkBold  &  &\CheckmarkBold       &$75.9$  &$78.5$  &$53.9$  &$70.1$  &$65.2$  &$72.5$  &$73.7$  &$78.4$  &$79.8$  &$37.7$  &$78.5$  &$68.8$  &$12.9$\\
\CheckmarkBold  &\CheckmarkBold  &\CheckmarkBold  &       &$75.7$  &$78.2$  &$54.7$  &$70.6$  &$65.6$  &$72.7$  &$73.6$  &$78.1$  &$79.7$  &\textbf{38.3}  &$79.4$  &$68.5$  &$12.6$\\
\CheckmarkBold  &\CheckmarkBold  &\CheckmarkBold  &\CheckmarkBold   &\textbf{76.1}  &\textbf{79.0}  &\textbf{56.6}  &\textbf{70.9}  &\textbf{65.9}  &\textbf{72.8}  &\textbf{74.0}  &\textbf{78.6}  &\textbf{80.0}  &\textbf{38.3}  &\textbf{79.5}  &\textbf{69.2}  &\textbf{13.3}\\\bottomrule
\end{tabular}
\vspace{-3mm}
\end{table*}

\begin{table}[t]
\centering
\setlength{\tabcolsep}{ 4.8pt}
\caption{Ablation study on the Anatomy-aware Knowledge Injection (AKI) module. We investigate the impact of feature hierarchy \textit{(last-layer feature v.s. intermediate feature)}, label format \textit{(one-hot label v.s. soft label)}, and anatomical granularity \textit{(grouped organs v.s. precised organs)}. The best results are \textbf{bolded}.}
\label{Table:ablation_aki}
\begin{tabular}{l|cc|cc|cc}
\toprule
\multirow{3}{*}{Settings}  &\multicolumn{2}{c|}{RadChestCT}  &\multicolumn{2}{c|}{MSD-Lung}  &\multicolumn{2}{c}{SegThor}\\ 
&\multicolumn{2}{c|}{(AUC)}  &\multicolumn{2}{c|}{(Dice)}  &\multicolumn{2}{c}{(Dice)}\\
&10\%  &100\%  &10\%  &100\%  &10\%  &100\%\\\midrule 
w/ Last-layer Feat.    &75.2  &77.1  &50.9  &69.9  &65.5  &72.1\\
w/ One-hot Label       &74.9  &77.4  &48.4  &68.1  &65.2  &71.9\\
w/ Grouped Organs      &76.0  &78.2  &48.7  &67.5  &64.9  &72.2\\\midrule
\algname               &\textbf{76.1}  &\textbf{79.0}  &\textbf{56.6}  &\textbf{70.9}  &\textbf{65.9}  &\textbf{72.8}\\\bottomrule
\end{tabular}
\vspace{-5mm}
\end{table}

\begin{table}[t]
\centering
\setlength{\tabcolsep}{ 4.5pt}
\caption{Ablation study on the Semantically-adaptive Selective Alignment (SSA) with different top-k value. The best results are \textbf{bolded}.}
\label{Table:ablation_ssa}
\begin{tabular}{l|cc|ccc|cc}
\toprule
\multirow{3}{*}{Settings}  &\multicolumn{2}{c|}{RadChestCT}  &\multicolumn{3}{c|}{STOIC-2021}  &\multicolumn{2}{c}{AHPH-10K}\\ 
&\multicolumn{2}{c|}{(AUC)}  &\multicolumn{3}{c|}{(AUC)}  &(AUC)  &(R@50)\\
&10\%  &100\%  &1\%  &10\%  &100\%  &\multicolumn{2}{c}{100\%}\\\midrule 
w/o            &73.1  &75.5  &72.2  &73.0  &77.4  &65.9  &11.6\\
w/ Top 16      &75.6  &78.0  &73.5  &74.3  &78.0  &68.0  &12.6\\
w/ Top 32      &75.8  &78.4  &74.2  &74.9  &78.7  &69.0  &13.0\\
w/ Top 64      &75.9  &78.1  &74.6  &75.1  &\textbf{79.3}  &69.1  &\textbf{13.5}\\
w/ Top 128     &76.3  &78.6  &72.5  &73.4  &77.8  &68.3  &12.5\\
w/ All         &76.0  &78.3  &69.7  &72.6  &79.0  &67.9  &12.0\\\midrule
\algname       &\textbf{76.1}  &\textbf{79.0}  &\textbf{75.0}  &\textbf{75.7}  &79.1  &\textbf{69.2}  &13.3\\\bottomrule
\end{tabular}
\vspace{-5mm}
\end{table}

\vspace{-3mm}
\subsection{Ablation studies}
We conduct ablation studies across representative downstream tasks, including classification, segmentation, prognosis prediction, and vision--language tasks (Table~\ref{Table:ablation}). 

\textbf{Anatomy-aware knowledge injection (AKI).} 
Introducing explicit anatomical priors consistently improves downstream performance, particularly for structure-sensitive tasks. 
Compared with the baseline (row 1 $\rightarrow$ row 2), AKI improves lesion segmentation on MSD-Lung by +3.3\% Dice and vocabulary classification on AHPH-10K by +3.4\% AUC. 
Removing AKI from the full model (row 8 $\rightarrow$ row 5) leads to clear performance degradation, especially for segmentation, demonstrating the effectiveness of anatomy-aware structural regularization.

\textbf{Semantically-adaptive selective alignment (SSA).} 
SSA consistently improves tasks requiring fine-grained cross-modal correspondence. 
Compared with the baseline (row 1 $\rightarrow$ row 3), SSA improves lesion segmentation by +4.7\% Dice and retrieval performance from 9.0\% to 11.8\% R@50, indicating more effective semantic grounding through adaptive region-text alignment.

\textbf{Semantically-adaptive fusion (Fusion).} 
Global fusion (GF) performs better on global semantic tasks such as classification, whereas semantic-localized fusion (SLF) is more effective for fine-grained dense prediction tasks. 
Combining both modules (row 8) consistently achieves the best overall performance, suggesting that global context aggregation and localized semantic modeling provide complementary benefits for unified volumetric representation learning.

\textbf{Specific module configurations.} 
Beyond component-level ablations, we further analyze key design choices within AKI and SSA especially under the 10\% low-data setting, where structural priors are particularly important.

\textit{(1) Configurations of Anatomy-aware Knowledge Injection (AKI):} 
Table~\ref{Table:ablation_aki} evaluates the effects of feature hierarchy, label format, and anatomical granularity. 
Injecting anatomical supervision into intermediate features consistently outperforms using final-layer features, suggesting that intermediate representations better preserve spatial anatomical structures. 
Replacing soft labels with one-hot labels degrades performance, indicating the importance of modeling partial organ occupancy for spatial supervision. 
Moreover, fine-grained organ partitioning performs better than grouped anatomical regions, demonstrating the benefit of more precise anatomical priors.

\textit{(2) Configurations of Semantically-adaptive Selective Alignment (SSA):}
Table~\ref{Table:ablation_ssa} evaluates different alignment scopes and selection strategies. 
Restricting alignment to top-$K$ regions consistently outperforms global alignment, indicating reduced background interference during cross-modal matching. 
Furthermore, dynamically selecting $K$ from predefined candidate sets achieves better performance than fixed-$K$ strategies, supporting our motivation that different clinical findings require different spatial grounding extents.

\begin{figure*}[t]
    \centering
    \includegraphics[width=\linewidth]{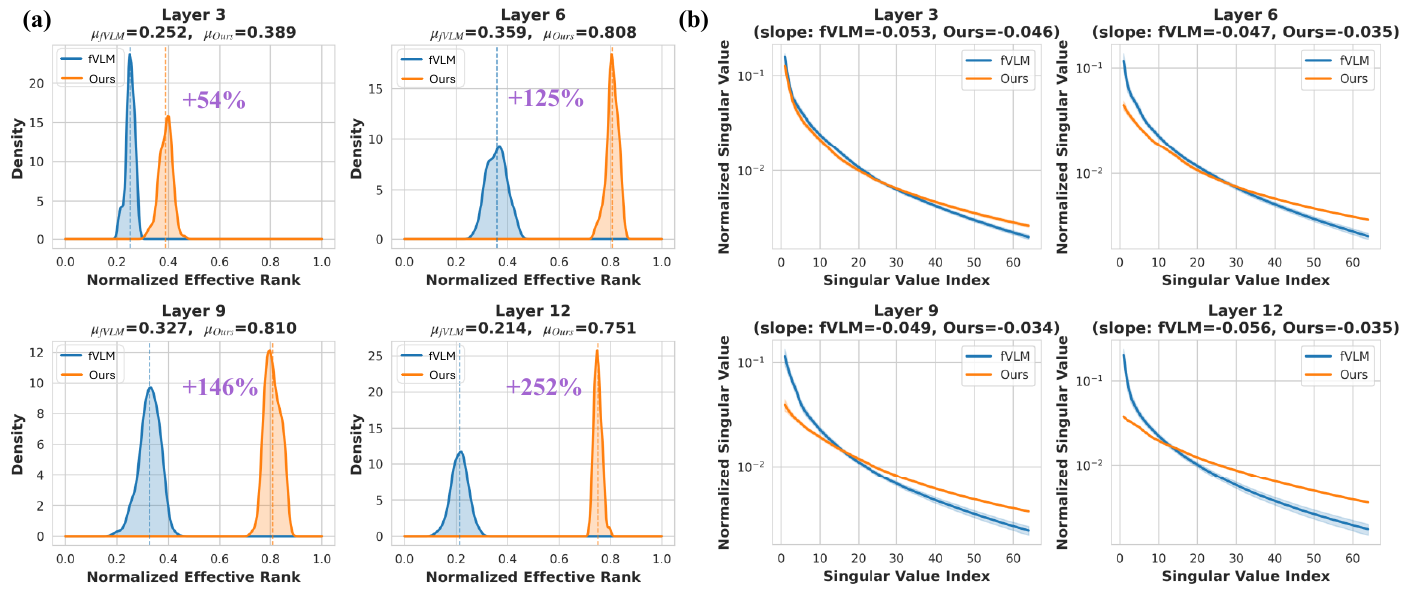}
    \caption{\textbf{Representation spectrum analysis across transformer layers}. 
(a) Distribution of normalized effective ranks computed from attention representations in different ViT layers. Higher effective rank indicates more diverse and less degenerated feature representations. Dashed vertical lines denote the mean effective rank of each model. 
(b) Mean singular value decay curves of attention representations on a logarithmic scale. \algname exhibits a slower spectral decay, suggesting that the learned representations preserve richer semantic variations across a broader range of feature directions.}
    \label{fig:efficent_rank}
    \vspace{-5mm}
\end{figure*}

\begin{figure}[t]
    \centering
    \includegraphics[width=\linewidth]{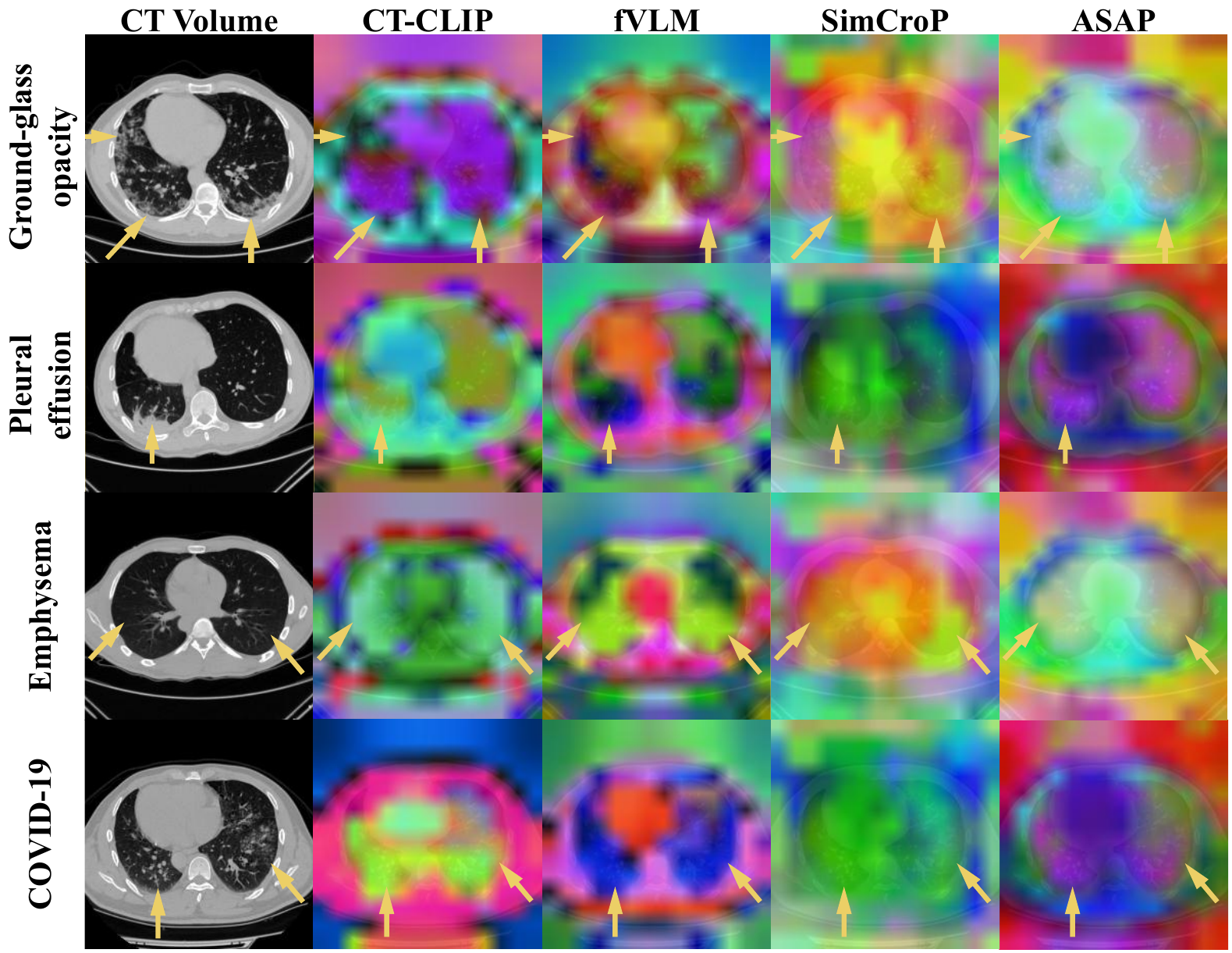}
    \caption{PCA visualizations of axial feature maps extracted directly from different pre-trained models on chest CT volumes. The \textcolor{yellowarrow}{arrows} indicate representative lesion regions. Compared with prior methods, \algname produces feature responses that exhibit clearer spatial localization and stronger semantic distinction around lesion-related regions.}
    \label{fig:pca}
    \vspace{-5mm}
\end{figure}

\vspace{-4mm}
\subsection{Representation Analysis}
To further analyze the learned representations, we examine the spectral properties of transformer attention features across different ViT layers using 500 randomly sampled CT volumes. Specifically, we compute the effective rank and singular value spectrum of intermediate attention representations. 
As shown in Fig.~\ref{fig:efficent_rank} (a), \algname consistently achieves higher effective ranks than fVLM, particularly in deeper transformer layers. Since effective rank reflects the diversity of dominant spectral components, these results suggest that \algname learns richer and less degenerated representations. 
Fig.~\ref{fig:efficent_rank} (b) further shows that \algname exhibits consistently slower singular value decay across middle and deep layers. This indicates that the learned feature energy is distributed across a broader range of spectral directions rather than concentrated in a few dominant modes, supporting improved representation diversity and semantic expressiveness.

\vspace{-3mm}
\subsection{Qualitative visualization}
PCA visualizations of axial feature maps extracted directly from different pre-trained models on chest CT volumes are presented in Fig.~\ref{fig:pca}.
We compare \algname with representative baseline methods~\cite{hamamci2024ctclip,fvlm_iclr25,wang2025simcrop} to analyze the spatial discriminability of learned volumetric representations. 
Compared with prior methods, the features extracted by \algname exhibit clearer activation patterns around lesion-related regions, such as opacity and pleural effusion. 
These observations suggest that \algname learns more semantically discriminative volumetric representations that better capture clinically relevant abnormal structures. 

\vspace{-3mm}
\section{Limitations and Future Directions}
\label{sec_limitation}
Although \algname demonstrates strong volumetric representation learning performance across comprehensive medical benchmarks, several limitations remain for future exploration:
\begin{itemize}

\item \textbf{Heterogeneous data distributions limit scalable cross-domain generalization}. 
Our \algname is primarily conducted on the CT-RATE dataset. 
Although we explored incorporating additional public CT volume--report datasets~\cite{huang2023inspect,vaya2020bimcv}, na\"{\i}vely combining heterogeneous data sources did not consistently improve downstream performance. 
We hypothesize that variations in report quality, terminology, annotation granularity, imaging protocols, and anatomical distributions hinder stable fine-grained vision--language alignment and cross-domain transferability. 
Future work will investigate large-scale multi-source pre-training with improved data harmonization and quality-aware alignment strategies.

\item \textbf{Representation learning alone may be insufficient for complex vision--language understanding tasks}. 
Although \algname consistently improves discriminative representation learning, the gains on report generation and certain VQA metrics remain comparatively limited. 
Besides visual representation quality, these tasks also rely on language modeling capacity, decoding strategies, and fine-grained multi-modal reasoning. 
Future research may therefore benefit from jointly optimizing volumetric representation learning with stronger generative and reasoning-oriented objectives.

\item \textbf{Model scaling and architectural extensibility remain underexplored}. 
Due to the limited scale of current volumetric vision--language datasets, increasing model capacity does not consistently yield improved performance. 
Moreover, the proposed masked modeling framework is currently developed based on ViT-style tokenized representations, and its extension to hierarchical architectures such as Swin Transformers has not yet been systematically studied. 
In addition, the relatively coarse input spacing adopted for computational efficiency may limit fine-grained anatomical modeling for dense prediction tasks compared with segmentation-oriented high-resolution pre-training methods~\cite{wu2025vocolarge,tang2026hiendmae}. 
Future work will explore scalable pre-training under larger data regimes, architecture-agnostic masked modeling strategies, and higher-resolution volumetric representation learning.

\item \textbf{Disease-aware knowledge injection remains unexplored}. 
The proposed anatomy-aware knowledge injection primarily models organ-level anatomical priors and does not explicitly incorporate fine-grained lesion-aware semantic supervision. 
Future advances in reliable volumetric lesion segmentation and disease localization tools may enable more precise disease-aware knowledge injection, potentially improving pathological representation learning and fine-grained clinical reasoning.
    
\end{itemize}
\vspace{-5mm}
\section{Conclusion}
In this paper, we propose \textbf{A}natomy-aware \textbf{S}emantically-\textbf{A}daptive \textbf{P}re-training (ASAP), a unified medical volumetric vision--language pre-training framework that explicitly models anatomy-aware structural priors and semantically-adaptive cross-modal alignment for chest CT representation learning. 
By addressing the semantic granularity mismatch between volumetric abnormalities and clinical descriptions, the proposed framework learns more transferable and data-efficient volumetric representations. 
Extensive experiments across classification, segmentation, prognosis prediction, and vision--language tasks demonstrate consistent improvements over prior Med-VLP methods, particularly under limited supervision. 
These findings highlight the importance of combining structured anatomical modeling with adaptive semantic grounding for scalable volumetric representation learning. 
Future work will explore large-scale multi-source pre-training, cross-modality generalization, and disease-aware semantic supervision for more generalizable medical vision--language foundation models.

\ifCLASSOPTIONcompsoc
  \section*{Acknowledgments}
\else
  \section*{Acknowledgment}
\fi
\noindent This work was supported by National Natural Science Foundation of China under Grant ( 62402473, 62271465), National Key R\&D Program of China under Grant 2025YFC3408300, Natural Science Foundation of Jiangsu Province (No. BK20255001), Suzhou Basic Research Program under Grant SYG202338, and Jiangsu Province Science Foundation for Youths (NO. BK20240464).


\bibliography{ref}

@String(CVPR  = {IEEE Conf. Comput. Vis. Pattern Recog.})

@String(ICCV  = {Int. Conf. Comput. Vis.})

@String(ECCV  = {Eur. Conf. Comput. Vis.})

@String(NeurIPS = {Adv. Neural Inform. Process. Syst.})

@String(ICML  = {Int. Conf. Mach. Learn.})

@String(ICLR  = {Int. Conf. Learn. Represent.})

@String(AAAI  = {AAAI})

@String(TMM   = {IEEE Trans. Multimedia})

@String(CVPR  = {CVPR})

@String(ICCV  = {ICCV})

@String(ECCV  = {ECCV})

@String(NeurIPS = {NeurIPS})

@String(ICML  = {ICML})

@String(ICLR  = {ICLR})

@String(TMM   =	{IEEE TMM})

@inproceedings{devlin2019bert,
  title={Bert: Pre-training of deep bidirectional transformers for language understanding},
  author={Devlin, Jacob and others},
  booktitle={NAACL},
  pages={4171--4186},
  year={2019}
}

@inproceedings{zhou2023mrm,
  title={Advancing Radiograph Representation Learning with Masked Record Modeling},
  author={Zhou, Hong-Yu and others},
  booktitle={ICLR},
  year={2023}
}

@inproceedings{clip,
  title={Learning transferable visual models from natural language supervision},
  author={Radford, Alec others},
  booktitle={ICML},
  pages={8748--8763},
  year={2021},
  organization={PMLR}
}

@article{coca,
  title={Coca: Contrastive captioners are image-text foundation models},
  author={Yu, Jiahui and others},
  journal={arXiv preprint arXiv:2205.01917},
  year={2022}
}

@inproceedings{huang2021gloria,
  title={Gloria: A multimodal global-local representation learning framework for label-efficient medical image recognition},
  author={Huang, Shih-Cheng and others},
  booktitle=ICCV,
  pages={3942--3951},
  year={2021}
}

@article{lu2019vilbert,
  title={Vilbert: Pretraining task-agnostic visiolinguistic representations for vision-and-language tasks},
  author={Lu, Jiasen and others},
  journal={NeurIPS},
  volume={32},
  year={2019}
}

@inproceedings{li2022flip,
  title={Scaling Language-Image Pre-training via Masking},
  author={Li, Yanghao and others},
  booktitle={CVPR},
  year={2023}
}

@inproceedings{vit,
  title={An image is worth 16x16 words: Transformers for image recognition at scale},
  author={Dosovitskiy, Alexey and others},
  booktitle={ICLR},
  year={2020}
}

@inproceedings{wang2022mgca,
  title={Multi-granularity cross-modal alignment for generalized medical visual representation learning},
  author={Wang, Fuying and Zhou, Yuyin and others},
  booktitle={NeurIPS},
  volume={35},
  pages={33536--33549},
  year={2022}
}

@inproceedings{zhang2022convirt,
  title={Contrastive learning of medical visual representations from paired images and text},
  author={Zhang, Yuhao and others},
  booktitle={Machine learning for healthcare conference},
  pages={2--25},
  year={2022},
  organization={PMLR}
}

@article{johnson2019mimic,
  title={MIMIC-CXR, a de-identified publicly available database of chest radiographs with free-text reports},
  author={Johnson, Alistair EW and others},
  journal={Scientific data},
  volume={6},
  number={1},
  pages={1--8},
  year={2019},
  publisher={Nature Publishing Group}
}

@inproceedings{su2020vlbert,
  title={VL-BERT: Pre-training of Generic Visual-Linguistic Representations},
  author={Su, Weijie and Zhu, Xizhou and Cao, Yue and Li, Bin and Lu, Lewei and Wei, Furu and Dai, Jifeng},
  booktitle={ICLR},
  year={2020}
}

@article{Oord2018RepresentationLW,
  title={Representation Learning with Contrastive Predictive Coding},
  author={A{\"a}ron van den Oord and Yazhe Li and Oriol Vinyals},
  journal={ArXiv},
  year={2018},
  volume={abs/1807.03748}
}

@inproceedings{He2021MAE,
  title={Masked Autoencoders Are Scalable Vision Learners},
  author={Kaiming He and others},
  booktitle=CVPR,
  year={2021},
  pages={15979-15988}
}

@article{zhou2021review,
  title={A review of deep learning in medical imaging: Imaging traits, technology trends, case studies with progress highlights, and future promises},
  author={Zhou, S Kevin and others},
  journal={Proceedings of the IEEE},
  year={2021},
  publisher={IEEE}
}

@article{tiu2022chexzero,
  title={Expert-level detection of pathologies from unannotated chest X-ray images via self-supervised learning},
  author={Tiu, Ekin and others},
  journal={Nature Biomedical Engineering},
  volume={6},
  number={12},
  pages={1399--1406},
  year={2022},
  publisher={Nature Publishing Group UK London}
}

@article{zhou2022referes,
  title={Generalized radiograph representation learning via cross-supervision between images and free-text radiology reports},
  author={Zhou, Hong-Yu and others},
  journal={Nature Machine Intelligence},
  volume={4},
  number={1},
  pages={32--40},
  year={2022},
  publisher={Nature Publishing Group UK London}
}

@inproceedings{wu2023medklip,
  title={MedKLIP: Medical Knowledge Enhanced Language-Image Pre-Training},
  author={Wu, Chaoyi and others},
  booktitle=ICCV,
  year={2023}
}

@article{zhang2023kad,
  title={Knowledge-enhanced visual-language pre-training on chest radiology images},
  author={Zhang, Xiaoman and Wu, Chaoyi and Zhang, Ya and Xie, Weidi and Wang, Yanfeng},
  journal={Nature Communications},
  volume={14},
  number={1},
  pages={4542},
  year={2023},
  publisher={Nature Publishing Group UK London}
}

@inproceedings{chen2022m3ae,
  title={Multi-modal masked autoencoders for medical vision-and-language pre-training},
  author={Chen, Zhihong and others},
  booktitle={MICCAI},
  pages={679--689},
  year={2022},
  organization={Springer}
}

@inproceedings{wan2023medUniC,
  title={Med-UniC: Unifying Cross-Lingual Medical Vision-Language Pre-Training by Diminishing Bias},
  author={Wan, Zhongwei and others},
  booktitle={NeurIPS},
  year={2023}
}

@inproceedings{cheng2023prior,
  title={PRIOR: Prototype Representation Joint Learning from Medical Images and Reports},
  author={Cheng, Pujin and others},
  booktitle=ICCV,
  pages={21361--21371},
  year={2023}
}

@inproceedings{li2022glip,
  title={Grounded language-image pre-training},
  author={Li, Liunian Harold and Zhang, Pengchuan and Zhang, Haotian and Yang, Jianwei and Li, Chunyuan and Zhong, Yiwu and Wang, Lijuan and Yuan, Lu and Zhang, Lei and Hwang, Jenq-Neng and others},
  booktitle={CVPR},
  pages={10965--10975},
  year={2022}
}

@article{pathak2022deep,
  title={Deep transfer learning based classification model for COVID-19 disease},
  author={Pathak, Yadunath and others},
  journal={Irbm},
  volume={43},
  number={2},
  pages={87--92},
  year={2022},
  publisher={Elsevier}
}

@article{liu2023medmllm,
  title={A medical multimodal large language model for future pandemics},
  author={Liu, Fenglin and others},
  journal={NPJ Digital Medicine},
  volume={6},
  number={1},
  pages={226},
  year={2023},
  publisher={Nature Publishing Group UK London}
}

@article{XIE2024MedIM,
title = {Rethinking masked image modeling for medical image representation},
journal = {MedIA},
pages = {103304},
year = {2024},
issn = {1361-8415},
doi = {https://doi.org/10.1016/j.media.2024.103304},
author = {Yutong Xie and others}
}

@inproceedings{li2024mlip,
  title={Mlip: Enhancing medical visual representation with divergence encoder and knowledge-guided contrastive learning},
  author={Li, Zhe and others},
  booktitle={CVPR},
  pages={11704--11714},
  year={2024}
}

@article{lin2024ctglip,
  title={CT-GLIP: 3D Grounded Language-Image Pretraining with CT Scans and Radiology Reports for Full-Body Scenarios},
  author={Lin, Jingyang and others},
  journal={arXiv preprint arXiv:2404.15272},
  year={2024}
}

@article{bai2024m3d,
  title={M3d: Advancing 3d medical image analysis with multi-modal large language models},
  author={Bai, Fan and others},
  journal={arXiv preprint arXiv:2404.00578},
  year={2024}
}

@article{hamamci2024ctclip,
  title={Generalist foundation models from a multimodal dataset for 3D computed tomography},
  author={Hamamci, Ibrahim Ethem and others},
  journal={Nature Biomedical Engineering},
  pages={1--19},
  year={2026},
  publisher={Nature Publishing Group UK London}
}

@inproceedings{fvlm_iclr25,
  title={Large-scale and fine-grained vision-language pre-training for enhanced CT image understanding},
  author={Shui, Zhongyi and others},
  booktitle={ICLR},
  pages={},
  year={2025}
}

@article{chen2022review,
  title={Recent advances and clinical applications of deep learning in medical image analysis},
  author={Chen, Xuxin and others},
  journal={MedIA},
  volume={79},
  pages={102444},
  year={2022},
  publisher={Elsevier}
}

@inproceedings{wang-etal-2022-medclip,
    title = "{M}ed{CLIP}: Contrastive Learning from Unpaired Medical Images and Text",
    author = "Wang, Zifeng  and
      others",
    booktitle = "EMNLP",
    month = dec,
    year = "2022",
    pages = "3876--3887"
}

@article{Wang2025ECAMP,
title = {ECAMP: Entity-centered Context-aware Medical Vision Language Pre-training},
journal = {MedIA},
volume = {105},
pages = {103690},
year = {2025},
issn = {1361-8415},
doi = {https://doi.org/10.1016/j.media.2025.103690},
author = {Rongsheng Wang and Qingsong Yao and Zihang Jiang and Haoran Lai and Zhiyang He and Xiaodong Tao and S. Kevin Zhou}
}

@inproceedings{he2024UniMedI,
  title={Unified Medical Image Pre-training in Language-Guided Common Semantic Space},
  author={He, Xiaoxuan and others},
  booktitle={ECCV},
  pages={123--139},
  year={2024},
  organization={Springer}
}

@article{ni2024mg3d,
title = {MG-3D: Multi-Grained Knowledge-Enhanced Vision-Language Pre-training for 3D Medical Image Analysis},
journal = {MedIA},
pages = {104027},
year = {2026},
issn = {1361-8415},
doi = {https://doi.org/10.1016/j.media.2026.104027},
author = {Xuefeng Ni and others},}

@inproceedings{hatamizadeh2022unetr,
  title={Unetr: Transformers for 3d medical image segmentation},
  author={Hatamizadeh, Ali and others},
  booktitle={WACV},
  pages={574--584},
  year={2022}
}

@inproceedings{cao2024biud,
  title={Bootstrapping chest ct image understanding by distilling knowledge from x-ray expert models},
  author={Cao, Weiwei and others},
  booktitle={CVPR},
  pages={11238--11247},
  year={2024}
}

@inproceedings{zhou2023self,
  title={Self pre-training with masked autoencoders for medical image classification and segmentation},
  author={Zhou, Lei and others},
  booktitle={2023 ISBI},
  pages={1--6},
  year={2023},
  organization={IEEE}
}

@inproceedings{landman2015btcv,
  title={Miccai multi-atlas labeling beyond the cranial vault--workshop and challenge},
  author={Landman, Bennett and others},
  booktitle={MICCAI Workshop},
  volume={5},
  pages={12},
  year={2015}
}

@inproceedings{boecking2022cxrbert,
  title={Making the Most of Text Semantics to Improve Biomedical Vision--Language Processing},
  author={Boecking, Benedikt and others},
  booktitle={ECCV},
  pages={1--21},
  year={2022}
}

@article{draelos2021radchestct,
  title={Machine-learning-based multiple abnormality prediction with large-scale chest computed tomography volumes},
  author={Draelos, Rachel Lea and others},
  journal={MedIA},
  volume={67},
  pages={101857},
  year={2021},
  publisher={Elsevier}
}

@article{he2020cc-ccii,
  title={Benchmarking deep learning models and automated model design for COVID-19 detection with chest CT scans},
  author={He, Xin and others},
  journal={MedRxiv},
  pages={2020--06},
  year={2020},
  publisher={Cold Spring Harbor Laboratory Press}
}

@article{setio2017luna16,
  title={Validation, comparison, and combination of algorithms for automatic detection of pulmonary nodules in computed tomography images: the LUNA16 challenge},
  author={Setio, Arnaud Arindra Adiyoso and others},
  journal={MedIA},
  volume={42},
  pages={1--13},
  year={2017},
  publisher={Elsevier}
}

@article{huang2024maco,
  title={Enhancing representation in radiography-reports foundation model: A granular alignment algorithm using masked contrastive learning},
  author={Huang, Weijian and others},
  journal={Nature Communications},
  volume={15},
  number={1},
  pages={7620},
  year={2024},
  publisher={Nature Publishing Group UK London}
}

@article{tang2026hiendmae,
  title={Hi-End-MAE: Hierarchical encoder-driven masked autoencoders are stronger vision learners for medical image segmentation},
  author={Tang, Fenghe and others},
  journal={MedIA},
  pages={103770},
  year={2026},
  publisher={Elsevier}
}

@inproceedings{wang2025simcrop,
  title={SimCroP: Radiograph Representation Learning with Similarity-Driven Cross-Granularity Pre-training},
  author={Wang, Rongsheng and others},
  booktitle={MICCAI},
  pages={563--573},
  year={2025},
  organization={Springer}
}

@InProceedings{Liu2025T3D,
    author    = {Liu, Che and others},
    title     = {T3D: Advancing 3D Medical Vision-Language Pre-training by Learning Multi-View Visual Consistency},
    booktitle = {ICCV Workshops},
    month     = {October},
    year      = {2025},
    pages     = {6704-6714}
}

@article{zhao2025hlip,
  title={Towards Scalable Language-Image Pre-training for 3D Medical Imaging},
  author={Chenhui Zhao and others},
  journal={Transactions on Machine Learning Research},
  issn={2835-8856},
  year={2026}
}

@article{wasserthal2023totalsegmentator,
  title={TotalSegmentator: robust segmentation of 104 anatomic structures in CT images},
  author={Wasserthal, Jakob and others},
  journal={Radiology: Artificial Intelligence},
  volume={5},
  number={5},
  pages={e230024},
  year={2023},
  publisher={Radiological Society of North America}
}

@inproceedings{cao2025vsd-boost,
  title={Boosting vision semantic density with anatomy normality modeling for medical vision-language pre-training},
  author={Cao, Weiwei and others},
  booktitle={ICCV},
  pages={23041--23050},
  year={2025}
}

@article{zhao2025sat,
  title={Large-vocabulary segmentation for medical images with text prompts},
  author={Zhao, Ziheng and others},
  journal={NPJ Digital Medicine},
  volume={8},
  number={1},
  pages={566},
  year={2025},
  publisher={Nature Publishing Group UK London}
}

@article{zhou2023pcrlv2,
  title={A unified visual information preservation framework for self-supervised pre-training in medical image analysis},
  author={Zhou, Hong-Yu and others},
  journal={TPAMI},
  volume={45},
  number={7},
  pages={8020--8035},
  year={2023},
  publisher={IEEE}
}

@ARTICLE{wu2025vocolarge,
  author={Wu, Linshan and Zhuang, Jiaxin and Chen, Hao},
  journal={TPAMI}, 
  title={Large-Scale 3D Medical Image Pre-Training With Geometric Context Priors}, 
  year={2025},
  volume={},
  number={},
  pages={1-18},
  doi={10.1109/TPAMI.2025.3639593}
}

@article{li2025visionunite,
  title={Visionunite: A vision-language foundation model for ophthalmology enhanced with clinical knowledge},
  author={Li, Zihan and others},
  journal={TPAMI},
  year={2025},
  publisher={IEEE}
}

@ARTICLE{xie2024unimiss+,
  author={Xie, Yutong and others},
  journal={TPAMI}, 
  title={UniMiSS+: Universal Medical Self-Supervised Learning From Cross-Dimensional Unpaired Data}, 
  year={2024},
  volume={46},
  number={12},
  pages={10021-10035},
  doi={10.1109/TPAMI.2024.3436105}
}

@ARTICLE{yan2026pamt,
  author={Yan, Rui and others},
  journal={TPAMI}, 
  title={Pathway-Aware Multimodal Transformer (PAMT): Integrating Pathological Image and Gene Expression for Interpretable Cancer Survival Analysis}, 
  year={2026},
  volume={48},
  number={1},
  pages={896-913},
  doi={10.1109/TPAMI.2025.3611531}
}

@ARTICLE{azad2024medsegreview,
  author={Azad, Reza and others},
  journal={TPAMI}, 
  title={Medical Image Segmentation Review: The Success of U-Net}, 
  year={2024},
  volume={46},
  number={12},
  pages={10076-10095},
  doi={10.1109/TPAMI.2024.3435571}
}

@ARTICLE{zhao2022diagnoselikeradiologist,
  author={Zhao, Gangming and others},
  journal={TPAMI}, 
  title={Diagnose Like a Radiologist: Hybrid Neuro-Probabilistic Reasoning for Attribute-Based Medical Image Diagnosis}, 
  year={2022},
  volume={44},
  number={11},
  pages={7400-7416},
  doi={10.1109/TPAMI.2021.3130759}
}

@ARTICLE{wu2023minimizingseg,
  author={Wu, Fuping and Zhuang, Xiahai},
  journal={TPAMI}, 
  title={Minimizing Estimated Risks on Unlabeled Data: A New Formulation for Semi-Supervised Medical Image Segmentation}, 
  year={2023},
  volume={45},
  number={5},
  pages={6021-6036},
  doi={10.1109/TPAMI.2022.3215186}
}

@ARTICLE{he2025homeomorphism,
  author={He, Yuting and others},
  journal={TPAMI}, 
  title={Homeomorphism Prior for False Positive and Negative Problem in Medical Image Dense Contrastive Representation Learning}, 
  year={2025},
  volume={47},
  number={5},
  pages={4122-4139},
  doi={10.1109/TPAMI.2025.3540644}
}

@ARTICLE{ma2022abdomenct,
  author={Ma, Jun and others},
  journal={TPAMI}, 
  title={AbdomenCT-1K: Is Abdominal Organ Segmentation a Solved Problem?}, 
  year={2022},
  volume={44},
  number={10},
  pages={6695-6714},
  doi={10.1109/TPAMI.2021.3100536}}

@article{wu2025radfm,
  title={Towards generalist foundation model for radiology by leveraging web-scale 2d\&3d medical data},
  author={Wu, Chaoyi and others},
  journal={Nature Communications},
  volume={16},
  number={1},
  pages={7866},
  year={2025},
  publisher={Nature Publishing Group UK London}
}

@article{zhang2024pmcvqa,
  title={Development of a large-scale medical visual question-answering dataset},
  author={Zhang, Xiaoman and others},
  journal={Communications Medicine},
  volume={4},
  number={1},
  pages={277},
  year={2024},
  publisher={Nature Publishing Group UK London}
}

@article{zheng2024large,
  title={Large-scale long-tailed disease diagnosis on radiology images},
  author={Zheng, Qiaoyu and others},
  journal={Nature Communications},
  volume={15},
  number={1},
  pages={10147},
  year={2024},
  publisher={Nature Publishing Group UK London}
}

@article{blankemeier2026merlin,
  title={Merlin: a computed tomography vision--language foundation model and dataset},
  author={Blankemeier, Louis and others},
  journal={Nature},
  pages={1--11},
  year={2026},
  publisher={Nature Publishing Group UK London}
}

@article{vaya2020bimcv,
  title={BIMCV COVID-19+: A large annotated dataset of RX and CT images from COVID-19 patients},
  author={Vay{\'a}, Maria De La Iglesia and others},
  journal={arXiv preprint arXiv:2006.01174},
  year={2020}
}

@article{lian2025alta,
  title={Efficient Medical Vision-Language Alignment Through Adapting Masked Vision Models},
  author={Lian, Chenyu and others},
  journal={TMI},
  year={2025},
  publisher={IEEE}
}

@InProceedings{Park_2025_radzero3d,
    author    = {Park, Jonggwon and others},
    title     = {RadZero3D: Bridging Self-Supervised Video Models and Medical Vision-Language Alignment for Zero-Shot Chest CT Interpretation},
    booktitle = {ICCV Workshops},
    month     = {October},
    year      = {2025},
    pages     = {6742-6749}
}

@inproceedings{li2023blipv2,
  title={Blip-2: Bootstrapping language-image pre-training with frozen image encoders and large language models},
  author={Li, Junnan and others},
  booktitle={ICML},
  pages={19730--19742},
  year={2023},
  organization={PMLR}
}

@inproceedings{zhai2023siglip,
  title={Sigmoid loss for language image pre-training},
  author={Zhai, Xiaohua and others},
  booktitle={ICCV},
  pages={11975--11986},
  year={2023}
}

@InProceedings{Singh2022flava,
    author    = {Singh, Amanpreet and others},
    title     = {FLAVA: A Foundational Language and Vision Alignment Model},
    booktitle = {CVPR},
    month     = {June},
    year      = {2022},
    pages     = {15638-15650}
}

@InProceedings{bao2021beit,
  title={Beit: Bert pre-training of image transformers},
  author={Bao, Hangbo and Dong, Li and Piao, Songhao and Wei, Furu},
  booktitle={ICLR},
  year={2022}
}

@inproceedings{wang2023beit3,
  title={Image as a foreign language: Beit pretraining for vision and vision-language tasks},
  author={Wang, Wenhui and others},
  booktitle={CVPR},
  pages={19175--19186},
  year={2023}
}

@ARTICLE{liu2025valor,
  author={Liu, Jing and others},
  journal={TPAMI}, 
  title={VALOR: Vision-Audio-Language Omni-Perception Pretraining Model and Dataset}, 
  year={2025},
  volume={47},
  number={2},
  pages={708-724},
  doi={10.1109/TPAMI.2024.3479776}
}

@ARTICLE{zhang2025unsupervised,
  author={Zhang, Dingwen and others},
  journal={TPAMI}, 
  title={Unsupervised Pre-Training With Language-Vision Prompts for Low-Data Instance Segmentation}, 
  year={2025},
  volume={47},
  number={10},
  pages={8642-8657},
  doi={10.1109/TPAMI.2025.3579469}
}

@inproceedings{jia2021ALIGN,
  title={Scaling up visual and vision-language representation learning with noisy text supervision},
  author={Jia, Chao and others},
  booktitle={ICML},
  pages={4904--4916},
  year={2021},
  organization={PMLR}
}

@article{liu2023llava,
  title={Visual instruction tuning},
  author={Liu, Haotian and Li, Chunyuan and Wu, Qingyang and Lee, Yong Jae},
  journal={NeurIPS},
  volume={36},
  pages={34892--34916},
  year={2023}
}

@article{alayrac2022flamingo,
  title={Flamingo: a visual language model for few-shot learning},
  author={Alayrac, Jean-Baptiste and others},
  journal={NeurIPS},
  volume={35},
  pages={23716--23736},
  year={2022}
}

@article{yao2025evax,
  title={Eva-x: A foundation model for general chest x-ray analysis with self-supervised learning},
  author={Yao, Jingfeng and others},
  journal={npj Digital Medicine},
  volume={8},
  number={1},
  pages={678},
  year={2025},
  publisher={Nature Publishing Group UK London}
}

@InProceedings{wu2024voco,
    author    = {Wu, Linshan and others},
    title     = {VoCo: A Simple-yet-Effective Volume Contrastive Learning Framework for 3D Medical Image Analysis},
    booktitle = {CVPR},
    month     = {June},
    year      = {2024},
    pages     = {22873-22882}
}

@inproceedings{li2025SAHN,
  title={Semantic-Aware Hard Negative Mining for Medical Vision-Language Contrastive Pretraining},
  author={Li, Yongxin and others},
  booktitle={Proceedings of the 33rd ACM International Conference on Multimedia},
  pages={3133--3142},
  year={2025}
}

@article{withers2021xray,
  title={X-ray computed tomography},
  author={Withers, Philip J and others},
  journal={Nature Reviews Methods Primers},
  volume={1},
  number={1},
  pages={18},
  year={2021},
  publisher={Nature Publishing Group UK London}
}

@article{liu2024Imitate,
  title={Imitate: Clinical prior guided hierarchical vision-language pre-training},
  author={Liu, Che and others},
  journal={TMI},
  year={2024},
  publisher={IEEE}
}

@article{liu2024g2d,
  title={G2d: From global to dense radiography representation learning via vision-language pre-training},
  author={Liu, Che and others},
  journal={NeurIPS},
  volume={37},
  pages={14751--14773},
  year={2024}
}

@inproceedings{he2023gsvl,
  title={Geometric visual similarity learning in 3d medical image self-supervised pre-training},
  author={He, Yuting and others},
  booktitle={CVPR},
  pages={9538--9547},
  year={2023}
}

@article{zhuang2025mim,
  title={Mim: Mask in mask self-supervised pre-training for 3d medical image analysis},
  author={Zhuang, Jiaxin and others},
  journal={TMI},
  year={2025},
  publisher={IEEE}
}

@article{li2025medvista3d,
  title={MedVista3D: Vision-Language Modeling for Reducing Diagnostic Errors in 3D CT Disease Detection, Understanding and Reporting},
  author={Li, Yuheng and others},
  journal={arXiv preprint arXiv:2509.03800},
  year={2025}
}

@article{niu2025m3fm,
  title={Medical multimodal multitask foundation model for lung cancer screening},
  author={Niu, Chuang and others},
  journal={Nature Communications},
  volume={16},
  number={1},
  pages={1523},
  year={2025},
  publisher={Nature Publishing Group UK London}
}

@article{revel2021stoic,
  title={Study of thoracic CT in COVID-19: the STOIC project},
  author={Revel, Marie-Pierre and others},
  journal={Radiology},
  volume={301},
  number={1},
  pages={E361--E370},
  year={2021},
  publisher={Radiological Society of North America}
}

@article{huang2023inspect,
  title={INSPECT: A Multimodal Dataset for Pulmonary Embolism Diagnosis and Prognosis},
  author={Huang, Shih-Cheng and others},
  journal={arXiv preprint arXiv:2311.10798},
  year={2023}
}

@article{antonelli2022msd,
  title={The medical segmentation decathlon},
  author={Antonelli, Michela others},
  journal={Nature communications},
  volume={13},
  number={1},
  pages={4128},
  year={2022},
  publisher={Nature Publishing Group UK London}
}

@article{roth2022covid-19-20,
  title={Rapid artificial intelligence solutions in a pandemic—The COVID-19-20 Lung CT Lesion Segmentation Challenge},
  author={Roth, Holger R and others},
  journal={MedIA},
  volume={82},
  pages={102605},
  year={2022},
  publisher={Elsevier}
}

@inproceedings{lambert2020segthor,
  title={Segthor: Segmentation of thoracic organs at risk in ct images},
  author={Lambert, Zo{\'e} and others},
  booktitle={IPTA},
  pages={1--6},
  year={2020},
  organization={Ieee}
}

@article{masoudi2018FUMPE,
  title={A new dataset of computed-tomography angiography images for computer-aided detection of pulmonary embolism},
  author={Masoudi, Mojtaba and others},
  journal={Scientific data},
  volume={5},
  number={1},
  pages={1--9},
  year={2018},
  publisher={Nature Publishing Group}
}

@article{bernard2018acdc,
  title={Deep learning techniques for automatic MRI cardiac multi-structures segmentation and diagnosis: is the problem solved?},
  author={Bernard, Olivier and others},
  journal={TMI},
  volume={37},
  number={11},
  pages={2514--2525},
  year={2018},
  publisher={ieee}
}

@article{tang2024ctrg,
  title={Work like a doctor: Unifying scan localizer and dynamic generator for automated computed tomography report generation},
  author={Tang, Yuhao and others},
  journal={Expert Systems with Applications},
  volume={237},
  pages={121442},
  year={2024},
  publisher={Elsevier}
}

@article{colak2021rspect,
  title={The RSNA pulmonary embolism CT dataset},
  author={Colak, Errol and others},
  journal={Radiology: Artificial Intelligence},
  volume={3},
  number={2},
  pages={e200254},
  year={2021},
  publisher={Radiological Society of North America}
}

@article{aerts2014rider,
  title={Decoding tumour phenotype by noninvasive imaging using a quantitative radiomics approach},
  author={Aerts, Hugo JWL and others},
  journal={Nature communications},
  volume={5},
  number={1},
  pages={4006},
  year={2014},
  publisher={Nature Publishing Group UK London}
}

@article{Ma2021COVID-19-SegBenchmark,
  title={Towards Data-Efficient Learning: A Benchmark for COVID-19 CT Lung and Infection Segmentation},
  author = {Ma, Jun and others},
  journal = {Medical Physics},
  volume = {48},
  number = {3},
  pages = {1197-1210},
  doi = {https://doi.org/10.1002/mp.14676},
  year = {2021}
}

@article{yang2023m2kt,
  title={Radiology report generation with a learned knowledge base and multi-modal alignment},
  author={Yang, Shuxin and others},
  journal={MedIA},
  volume={86},
  pages={102798},
  year={2023},
  publisher={Elsevier}
}

@article{Qwen2.5-VL,
  title={Qwen2.5-VL Technical Report},
  author={Bai, Shuai and others},
  journal={arXiv preprint arXiv:2502.13923},
  year={2025}
}

@article{zhang2025radgenomect,
  title={Development of a large-scale grounded vision language dataset for chest CT analysis},
  author={Zhang, Xiaoman and others},
  journal={Scientific Data},
  volume={12},
  number={1},
  pages={1636},
  year={2025},
  publisher={Nature Publishing Group UK London}
}

@article{yang2025qwen3,
  title={Qwen3 technical report},
  author={Yang, An and others},
  journal={arXiv preprint arXiv:2505.09388},
  year={2025}
}

@inproceedings{liu2023m3ae,
  title={M3AE: multimodal representation learning for brain tumor segmentation with missing modalities},
  author={Liu, Hong and others},
  booktitle={AAAI},
  volume={37},
  number={2},
  pages={1657--1665},
  year={2023}
}

@article{huang2024enhancing,
  title={Enhancing the vision--language foundation model with key semantic knowledge-emphasized report refinement},
  author={Huang, Weijian and others},
  journal={MedIA},
  volume={97},
  pages={103299},
  year={2024},
  publisher={Elsevier}
}

@article{yan2025mgvla,
  title={Multi-Grained Vision-and-Language Model for Medical Image and Text Alignment},
  author={Yan, Huimin and others},
  journal={TMM},
  year={2025},
  publisher={IEEE}
}

@article{chen2024huatuogptvision,
      title={HuatuoGPT-Vision, Towards Injecting Medical Visual Knowledge into Multimodal LLMs at Scale}, 
      author={Junying Chen and others},
      year={2024},
      journal={arXiv preprint arXiv:2406.19280},
      primaryClass={cs.CV},
}

@ARTICLE{yu2025med3dvlm,
  author={Xin, Yu and others},
  journal={JBHI}, 
  title={Med3DVLM: An Efficient Vision-Language Model for 3D Medical Image Analysis}, 
  year={2026},
  volume={30},
  number={3},
  pages={2524-2536},
  doi={10.1109/JBHI.2025.3604595}}

@inproceedings{zhang2020bertscore,
  author       = {Tianyi Zhang and others},
  title        = {BERTScore: Evaluating Text Generation with {BERT}},
  booktitle    = {ICLR},
  year         = {2020}
}
\bibliographystyle{IEEEtran}

\end{document}